\pdfoutput=1
\documentclass[11pt]{article}

\usepackage{EMNLP2023}

\usepackage{times}
\usepackage{latexsym}
\usepackage[T1]{fontenc}
\usepackage[utf8]{inputenc}

\usepackage{microtype}

\usepackage{inconsolata}

\usepackage{CJKutf8}

\usepackage{hyperref}
\usepackage{url}

\usepackage{amssymb}
\usepackage{amsmath}
\usepackage{amsthm}
\usepackage{amsfonts}
\usepackage{bm}
\usepackage{nicefrac}
\usepackage{dsfont}
\usepackage{cleveref}
\usepackage{fancyvrb}
\usepackage{fvextra}
\usepackage{listings}
\usepackage{xcolor}
\usepackage{tabularx}% added for table design
\usepackage[all]{nowidow}
\usepackage[inline]{enumitem}
\usepackage{nicefrac}       % compact symbols for 1/2, etc.
\usepackage{microtype}      % microtypography
\usepackage{xcolor}         % colors
\usepackage{booktabs}
\usepackage{multirow}
\usepackage{graphicx}
\usepackage{subcaption}
\usepackage{caption}
\usepackage{rotating}
\usepackage[normalem]{ulem}
\usepackage{colortbl}
\usepackage{linguex}

\usepackage{booktabs}
\usepackage{multirow}
\usepackage{makecell}
\usepackage{arydshln}

\usepackage{color}
\usepackage{xcolor}
\usepackage{colortbl}

\usepackage{comment}
\usepackage{xspace}
\usepackage{soul}

\usepackage{enumitem}
\usepackage[normalem]{ulem}
\setlist[itemize,enumerate]{leftmargin=*}
\usepackage{pifont}% http://ctan.org/pkg/pifont

\makeatletter
\def\adl@drawiv#1#2#3{%
        \hskip.5\tabcolsep
        \xleaders#3{#2.5\@tempdimb #1{1}#2.5\@tempdimb}%
                #2\z@ plus1fil minus1fil\relax
        \hskip.5\tabcolsep}
\newcommand{\cdashlinelr}[1]{%
  \noalign{\vskip 2pt
           \global\let\@dashdrawstore\adl@draw
           \global\let\adl@draw\adl@drawiv}
  \cdashline{#1}[.4pt/2pt]
  \noalign{\global\let\adl@draw\@dashdrawstore
           \vskip 2pt}}
\makeatother

\usepackage{tikz,pgfplots,pgfplotstable}

\usepackage{siunitx} 
\usepackage{multirow, makecell}
\usepackage{pifont}% http://ctan.org/pkg/pifont
\newcommand{\cmark}{\ding{51}}%
\newcommand{\xmark}{\ding{55}}%
\usepackage{siunitx} 
\usepackage{multirow, makecell}
\usepackage{tabularx}% added for table design
\usepackage[all]{nowidow}
\definecolor{darkgreen}{rgb}{0.0, 0.42, 0.24}
\definecolor{green}{RGB}{112, 173,71}
\definecolor{blue}{RGB}{68, 114,196}
\definecolor{orange}{RGB}{237, 125,49}
\definecolor{red}{RGB}{202, 54,49}
\definecolor{yellow}{RGB}{222,194, 142}
\newcommand{\chrf}{\textsc{chrF}\xspace}
\newcommand{\bleu}{\textsc{BLEU}\xspace}
\newcommand{\comet}{\textsc{Comet}\xspace}
\newcommand{\muda}{\textsc{MuDA}\xspace}

\newcommand{\contextqe}{\textsc{ContextCometQE}\xspace}
\newcommand{\contextmqm}{\textsc{ContextMQM}\xspace}

\newcommand{\bryan}[1]{{\textcolor{red}{[BE]}}}

\definecolor{darkblue}{rgb}{0,0,.5}
\definecolor{darkgreen}{rgb}{0,.5,0}
\definecolor{lightgray}{rgb}{.8,.8,.8}
\definecolor{aliceblue}{rgb}{0.75, 0.75, 1.0}
\definecolor{darkseagreen}{rgb}{0.46, 0.74, 0.46}
\definecolor{alizarin}{rgb}{0.82, 0.1, 0.26}
\definecolor{airforceblue}{rgb}{0.36, 0.54, 0.66}
\definecolor{red_graph}{rgb}{0.98, 0.8, 0.8}
\definecolor{blue_graph}{rgb}{0.8, 0.98, 0.8}
\definecolor{red}{rgb}{0.8, 0.0, 0.0}

\newcommand{\deeptext}{\textsc{DeepText Lab}\xspace}
\newcommand{\huwaei}{\textsc{HW-TSC}\xspace}
\newcommand{\multitan}{\textsc{Multitan-GML}\xspace}
\newcommand{\adapt}{\textsc{SETU-ADAPT}\xspace}
\newcommand{\sheffield}{\textsc{SheffieldGATE}\xspace}
\newcommand{\clteam}{\textsc{CLTeam}\xspace}
\newcommand{\dcu}{\textsc{DCUGenNLP}\xspace}
\newcommand{\unbabel}{\textsc{Unbabel-IT}\xspace}
\newcommand{\baseline}{\textsc{NLLB-3.3}B\xspace}

\newcommand{\fr}{\textsc{en-fr}\xspace}
\newcommand{\de}{\textsc{en-de}\xspace}
\newcommand{\pt}{\textsc{en-pt}\xspace}
\newcommand{\nl}{\textsc{en-nl}\xspace}
\newcommand{\ko}{\textsc{en-ko}\xspace}

\usepackage{fontawesome}
\title{Findings of the WMT 2024 Shared Task on Chat Translation}

\author{
    Wafaa Mohammed$^{1}$ \quad
    Sweta Agrawal$^{2}$  \quad 
    M. Amin Farajian$^{3}$ \\
    \bf
    Vera Cabarrão$^{3}$ \quad
    Bryan Eikema$^{1}$ \quad
    Ana C. Farinha$^{3}$ \quad
    José G. C. de Souza$^{3}$ \quad 
    \\
    \smallskip
    \\
    $^{1}$University of Amsterdam, Netherlands \\
    $^{2}$Instituto de Telecomunicações, Lisbon, Portugal \\
    $^{3}$Unbabel, Lisbon, Portugal \\
}

\begin{document}
\maketitle
\begin{abstract}
This paper presents the findings from the third edition of the Chat Translation Shared Task. As with previous editions, the task involved translating bilingual customer support conversations, specifically focusing on the impact of conversation context in translation quality and evaluation. We also include two new language pairs: English$\leftrightarrow$Korean and English$\leftrightarrow$Dutch, in addition to the set of language pairs from previous editions: English$\leftrightarrow$German, English$\leftrightarrow$French, and English$\leftrightarrow$Brazilian Portuguese.

We received 22 primary submissions and 32 contrastive submissions from eight teams, with each language pair having participation from at least three teams.  We evaluated the systems comprehensively using both automatic metrics and human judgments via a direct assessment framework.
The official rankings for each language pair were determined based on human evaluation scores, considering performance in both translation directions—agent and customer. Our analysis shows that while the systems excelled at translating individual turns, there is room for improvement in overall conversation-level translation quality.

\end{abstract}

\section{Introduction}
Translating conversational text, in particular customer support chats, is an important and challenging application for machine translation (MT) technology. According to a 2020 survey from CSA Research, 75\% of shoppers are more likely to make another purchase if customer support is offered in their native language, making it appealing for businesses to invest in multilingual support.\footnote{\url{https://csa-research.com/Featured-Content/For-Global-Enterprises/Global-Growth/CRWB-Series/CRWB-B2C}} However, there are several key challenges to translating chats:
customer support chats typically feature short text exchanges between agents and customers (see Table~\ref{table:conv-example}), leading to fragmented sentences and omission of information (implied by the context). This makes it difficult for MT systems to produce coherent translations that maintain the intended meaning of the text \cite{farajian-etal-2020-findings}. Furthermore, chats often use colloquial language and are characterized by informality and grammatical inaccuracies \cite{goncalves-etal-2022-agent}. Consequently, translating such content poses a dual challenge: not only must a system accurately translate between languages, but it should also effectively model the nuances and ambiguity in a dialogue. 

\begin{table*}[t]
\small
\centering
\scalebox{0.8}{
\begin{tabular}{ll}
\toprule
\multirow{2}{*}{\faUser \ customer}  & {\fontfamily{cmss}\selectfont Hallo, ich komme nicht in meine Sum up pos was denn no App rein} \\
 & \textcolor{black!40}{{\fontfamily{cmss}\selectfont Hello, I can not get into my sum up pos what then no app}}  \\
\midrule
\multirow{2}{*}{\faHeadphones \ agent} & {\fontfamily{cmss}\selectfont I am sorry to hear that.}  \\
\multicolumn{1}{l}{}  & \textcolor{black!40}{{\fontfamily{cmss}\selectfont Es tut mir leid, das zu erfahren.}} \\
\midrule
\multirow{2}{*}{\faHeadphones \ agent} & {\fontfamily{cmss}\selectfont Let me see what I can do for you} \\
 & \textcolor{black!40}{{\fontfamily{cmss}\selectfont Lassen Sie mich sehen, was ich für Sie tun kann.}} \\
 \midrule
\multirow{2}{*}{\faHeadphones \ agent}  & {\fontfamily{cmss}\selectfont Could you please tell me what error message you can see while logging in to your POS?} \\
 & \textcolor{black!40}{{\fontfamily{cmss}\selectfont Könnten Sie mir bitte sagen, welche Fehlermeldung Sie sehen können, während Sie sich bei Ihrem POS anmelden?}} \\
\midrule
\multirow{2}{*}{\faUser \ customer}  & {\fontfamily{cmss}\selectfont Wenn ich auf die App gehe, erscheint dieses Gerät hinzufügen.} \\
 & \textcolor{black!40}{{\fontfamily{cmss}\selectfont When I go to the app, it shows Add this device.}} \\
 \midrule
\multirow{2}{*}{\faHeadphones \ agent}  &  {\fontfamily{cmss}\selectfont Could you please try to connect the App with the POS?} \\
 & \textcolor{black!40}{{\fontfamily{cmss}\selectfont Könnten Sie bitte versuchen, die App mit dem POS zu verbinden?}} \\
\midrule
\multirow{2}{*}{\faUser \ customer}  & {\fontfamily{cmss}\selectfont die App ist die PRS-ORG pos app} \\
 & \textcolor{black!40}{{\fontfamily{cmss}\selectfont the app is the PRS-ORG app}}   \\
 \midrule
\multirow{2}{*}{\faUser \ customer}  &  {\fontfamily{cmss}\selectfont ich habe die Frage daher nicht verstanden} \\
 & \textcolor{black!40}{{\fontfamily{cmss}\selectfont so I did not understand the question}} \\
 \midrule
\multirow{2}{*}{\faHeadphones \ agent}  & {\fontfamily{cmss}\selectfont Could you please elaborate on your query?} \\
 & \textcolor{black!40}{{\fontfamily{cmss}\selectfont Könnten Sie bitte Ihre Anfrage näher erläutern?}} \\

\bottomrule
\end{tabular}
}
\caption{An example of a \de conversation between a \textit{customer} (\faUser) and an \textit{agent} (\faHeadphones) from MAIA dataset.}
\label{table:conv-example}
\end{table*}

While recent advancements in MT systems, driven by LLMs, have proven effective in various tasks, bilingual chat translation remains under-explored. 
The \textbf{Chat Translation Shared Task} aims to bridge this gap by promoting research and development of MT systems designed specifically for conversational translation. This year's edition places special emphasis on the role of conversation context, encouraging teams to examine how context influences translation in the inherently ambiguous and dynamic nature of chat interactions.
Following the success of the previous two editions of the Chat
Translation Shared Task \citep{farajian-etal-2020-findings,farinha-etal-2022-findings}, this year we organized the third edition of the task with the following improvements:
\begin{itemize}
  \setlength{\itemsep}{2pt}
    \item We expanded the set of language pairs to include English$\leftrightarrow$Korean (\ko) and English$\leftrightarrow$Dutch (\nl), in addition to languages from previous editions:
    English$\leftrightarrow$German (\de), English$\leftrightarrow$French (\fr), and English$\leftrightarrow$Brazilian Portuguese (\pt).
    \item We carefully curated the evaluation sets to enable the evaluation of effective context utilization on systems' performance.
     \item We conducted a comprehensive evaluation of all systems using: a) automatic metrics (both neural and lexical) that assess translation quality and the accuracy of modeling discourse phenomena using \muda \cite{fernandes-etal-2023-translation}, b) human direct assessments by professional linguists, and c) LLM-based fine-grained error analysis following the MQM framework.
\end{itemize}

\noindent
We received a total of 22 primary submissions, 6 submissions for en$\leftrightarrow$de, 5 for en$\leftrightarrow$fr, 4 for en$\leftrightarrow$nl, 4 for en$\leftrightarrow$pt-br, and 3 for en$\leftrightarrow$ko. Six out of the eight teams used large language models (LLMs) as their base translation model, implementing various strategies such as finetuning on shared task data, augmenting training data with synthetic datasets, prompting strategies, quality-aware decoding, and several ways of leveraging conversational context to improve translation quality. With these multi-faceted solutions explored by several teams, this year's shared task yields valuable insights into the effectiveness of LLMs in translating conversational texts. We summarize the key findings from the shared task below:

\begin{itemize}[leftmargin=*]
    \item Incorporating contextual information from previous turns almost always improved translation quality. However, the optimal method for introducing context (whether through summary, graph, or raw context) still requires further investigation.
    \item Human evaluation showed that turn-level translation quality was consistently high across all participating systems and language pairs. Nonetheless, there is room for improvement in translating texts from later turns and at the conversation level as a whole.
    \item The \unbabel submission achieved the best results across most language pairs and evaluation criteria, except on the \de and \fr tasks according to automatic metrics. 
\end{itemize}

These findings suggest that future editions of the shared task could benefit from a) designing evaluation frameworks, both automatic and human, that specifically target dialogue-specific criteria to better understand system limitations \cite{yeh-etal-2021-comprehensive, a-2022-automating, deriu2021survey}; b) expanding the datasets to include more challenging domains (e.g. patient-physician conversation or everyday dialogues) and contexts (e.g. multimodal chats)  for a more thorough evaluation of MT systems.

\section{Task Description}
\label{sec:task_description}
As in previous editions of the task, we evaluate the effectiveness of a translation layer in translating text from the customer’s language to the agent’s language (e.g., English) and vice versa. We provide real bilingual customer support data for five different language pairs and encourage the participants to use conversation context. They are asked to submit translations for both directions (agent and customer). We detail the shared task dataset provided to the participants and evaluation in \S~\ref{sec:maia_v2} and \S~\ref{sec:evaluation} respectively.

\subsection{Data: The MAIA 2.0 Corpus}
\label{sec:maia_v2}

\begin{table*}[]
    \centering
    \renewcommand{\arraystretch}{1.1}
    \footnotesize
    \resizebox{\linewidth}{!}{
   \begin{tabular}{lrrrrrrrrrrrr}
   \toprule
    & \multicolumn{4}{c}{\textbf{train}}  &\multicolumn{4}{c}{\textbf{dev}} & \multicolumn{4}{c}{\textbf{test}} \\
  % \midrule
  \cline{2-13}
  \addlinespace[0.1cm]
 \multirow{-2}{*}{\textbf{LP}} & \multicolumn{1}{c}{\# seg} &  \multicolumn{1}{c}{\# conv} &  \multicolumn{1}{c}{\# length} &  \multicolumn{1}{c}{\# words} & \multicolumn{1}{c}{\# seg} &  \multicolumn{1}{c}{\# conv} &  \multicolumn{1}{c}{\# length} &  \multicolumn{1}{c}{\# words} & \multicolumn{1}{c}{\# seg} &  \multicolumn{1}{c}{\# conv} &  \multicolumn{1}{c}{\# length} &  \multicolumn{1}{c}{\#words} \\
  \midrule
\nl & 15.5k & 595 & 26.0 & 8.6 & 2.5k & 72 & 35.4 & 9.8 & 2k & 58 & 34.7 & 10.2 \\
\pt & 15.0k & 435 & 34.7 & 8.0 & 2.5k & 96 & 26.6 & 8.8 & 2k & 73 & 27.9 & 8.8 \\
\de	& 17.8k & 493 & 36.1 & 8.5 & 2.5k & 82 & 31.3 & 9.4 & 2k & 67 & 30.5 & 9.4 \\
\ko	& 16.1k & 423 & 38.1 & 8.5 & 1.8k & 38 & 50.9 & 10.5 & 2k & 42 & 47.2 & 9.6 \\
\fr	& 15.0k & 264 & 56.9 & 7.7 & 3.0k & 90 & 33.4 & 10.1 & 2k & 65 & 32.2 & 10.1 \\
\bottomrule
\end{tabular}
}
    \caption{Dataset statistics with the number of segments (\#seg), number of conversations (\#conv), average conversation length (\#length), and average number of words per turn (\#words) in each split. Note that for \textsc{ko} customer parts, we considered the English reference translation to calculate the number of words. 
    %\wafaa{can someone handle the columns alignment and spacing?}\jose{it seems ok to me, is something off?}
    }
    \label{tab:data_stats}
\end{table*}

The MAIA 2.0 corpus builds upon the dataset released in the previous edition \citep{farinha-etal-2022-findings} and includes two additional language pairs: Dutch and Korean. Furthermore, we expanded the sizes of the existing language pairs, ensuring that each language pair contained approximately 20k segments. The dataset encompasses dialogues across diverse topics, including account registration issues, payment and delivery clarifications, and after-sale services in various industries such as retail and gaming. The new dataset was automatically anonymized using Unbabel’s proprietary anonymization tool, followed by a manual validation performed by expert linguists, to comply with the General Data Protection Regulation (GDPR). The corpus is released under the CC-BY-NC-4.0 license and can be freely used for research purposes only. Please note that, as the license states, no commercial uses are permitted for this corpus. 

\paragraph{Training and Evaluation Datasets.} We provide both training and evaluation (development and test) sets that participants can use to build their systems. \Cref{tab:data_stats} presents each data splits' statistics, including the number of segments, conversations, and average conversation length. We construct the development and test sets by selecting conversations that exhibit the highest counts of context-dependent discourse phenomena tags, as extracted using Multilingual Discourse Aware (\muda) tagger \cite{fernandes-etal-2023-translation}.

\subsection{Evaluation}
\label{sec:evaluation}

We perform a comprehensive evaluation of all submitted systems, using both automatic and human evaluation. Official rankings are determined based on the human assessment scores for both customer and agent translations. We outline the various evaluations conducted below:

\subsubsection{Automatic Evaluation}
\label{subsec:automatic_evaluation}
We use \comet \cite{rei-etal-2022-comet} as our primary evaluation metric for assessing translation quality of the submitted systems.\footnote{\href{https://huggingface.co/Unbabel/wmt22-comet-da}{\texttt{Unbabel/wmt22-comet-da}}} Additionally, we report lexical metrics: \bleu and \chrf using the SacreBLEU library \cite{post-2018-call}. We also include \contextqe  \citep{DBLP:journals/corr/abs-2403-08314}, a reference-free metric that uses bilingual context (previous two turns) to assess the translation quality of the current turn. As efficient discourse handling is not directly reflected in standard MT metrics (both lexical and neural), we report the F1 accuracy on the \muda-tagged discourse phenomena.
We considered 4 context-dependent discourse phenomena in our analysis:
\begin{itemize}
    \item \textbf{Lexical cohesion:} Entities may have multiple possible translations in the target language, but the same entity should be referred to by the same word in a conversation.
    \item \textbf{Formality:} Korean uses honorifics to indicate formality, which are special titles or words expressing courtesy or respect for position. In other languages, speakers use second-person pronouns to refer to someone more formally or informally, depending on their relationship with the addressee. Formality should be consistent throughout a conversation. 
    \item \textbf{Pronoun resolution:} Some highly inflected languages use gendered pronouns based on semantic or morphological rules. 
    To assign the correct pronoun, it is therefore necessary to use the conversation's context to distinguish the grammatical gender of the pronoun's antecedent.
    \item \textbf{Verb forms:} Verbs must be translated consistently using the form that reflects the tone, and mood of both parties in the conversation.
\end{itemize}

\begin{figure*}[t]
    \centering
    \includegraphics[width=0.99\linewidth]{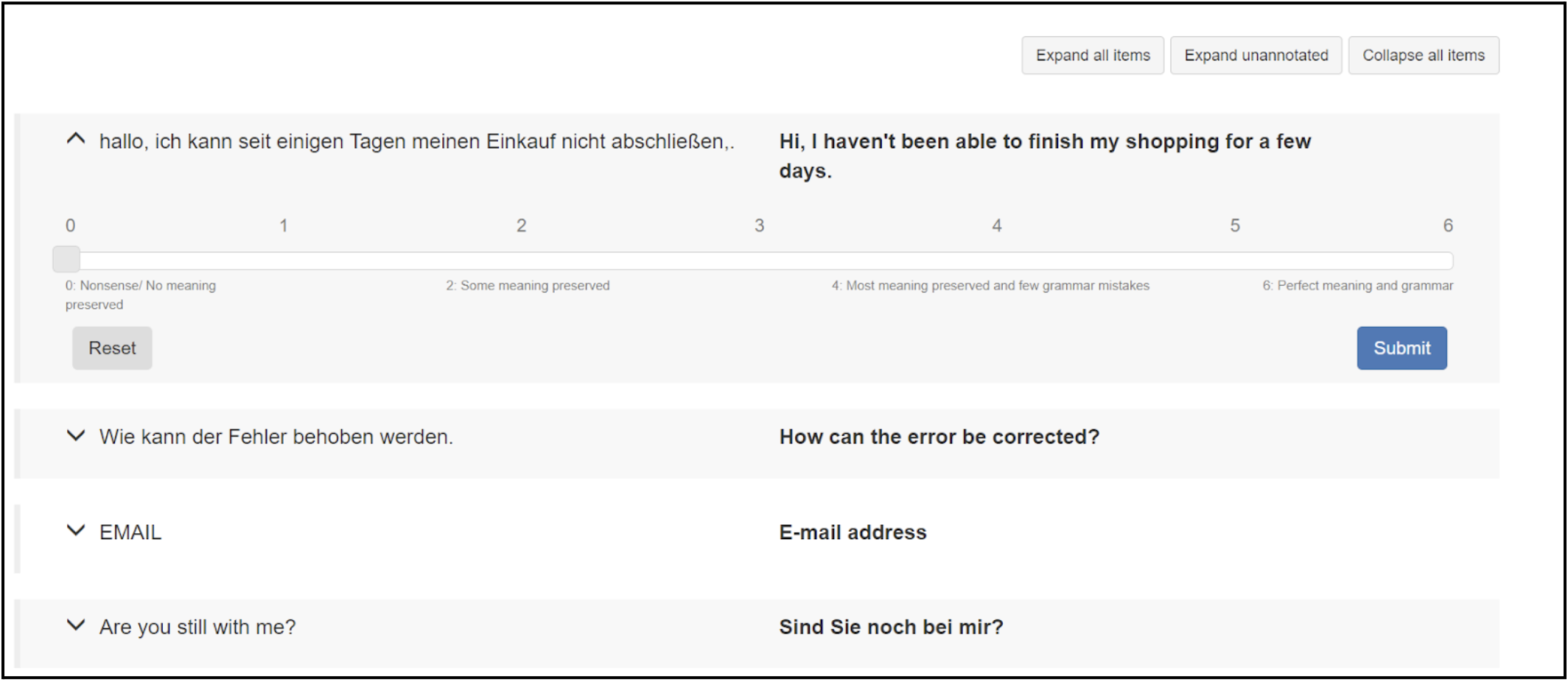}
    \caption{Screen capture of the Appraise interface used by professional linguists to perform human evaluation.}
    \label{fig:appraise}
\end{figure*}

\begin{table}[t]
    \centering
    \small
    \scalebox{0.9}{
   \begin{tabular}{lrrrr}
   \toprule
  \textbf{LP}  & \textbf{Threshold} & \textbf{\# Chats} & \textbf{\# Systems} & \textbf{\# annotated} \\
  & & & & \textbf{segments} \\
  
  \midrule
    \textsc{en-nl} &	35	& 27 &	5 &	3830 \\
\textsc{en-pt}&	28 &	41&	5	&	4700 \\
\textsc{en-de}	&	31	&36& 7	&	6629 \\
\textsc{en-ko}	&	48&	24&	4&		3648 \\
\textsc{en-fr}	&	33&	37&	6	&	6324 \\
\bottomrule
\end{tabular}}
    \caption{Statistics of the conversations and instances sampled for the human evaluation step.}
    \label{tab:human_data}
    \vspace{-0.3cm}
\end{table}

\subsubsection{Manual Evaluation}
\label{subsec:manual_evaluation}
We use the DA+SQM (Direct Assessment + Scalar Quality Metric) evaluation framework, following the campaigns conducted by the WMT General Translation track over the past years, implemented via the Appraise framework \cite{federmann-2018-appraise} to collect human assessments of translation quality for the submitted systems. We ask professional linguists hired via the UpWork\footnote{\url{upwork.com}} platform to evaluate each turn in a conversation within the full context and provide a conversation-level quality score on a continuous scale from 0 to 100. They were instructed to pay special attention to conversation-level properties such as the consistency of style, selection of terms, formality, etc in addition to the correctness criteria. The quality scale includes seven labeled tick marks representing various quality levels based on both accuracy and grammatical correctness (Figure \ref{fig:appraise}).

\paragraph{Data Selection} 
For the human evaluation, we retain conversations with up to a given number of turns to make the evaluation manageable. The number of turns for each language pair is specified in Table \ref{tab:human_data} (``Threshold''), together with the number of conversations and instances retained.

\paragraph{Measure} We generate turn-level and conversation-level system rankings for each language pair by aggregating the direct assessment scores provided by the linguists at the turn level and the conversation level respectively.

\subsubsection{LLM-based Error Assessments}

LLM-based evaluation has garnered a lot of interest from the community for conducting human-like evaluations. This shift is largely driven by the increasing complexity and scale of language models, making them capable of capturing nuanced understanding and performance of models in real-world tasks. For MT, LLM-based metrics are used to provide fine-grained error assessments over the nature, type, and severity of the errors following the MQM framework \cite{fernandes-etal-2023-devil, lu-etal-2024-error, kocmi-federmann-2023-gemba}. Recently, \citet{DBLP:journals/corr/abs-2403-08314} show that context-aware prompting for deriving MQM assessment using LLMs can achieve better correlation with human judgments than the standard MQM prompt for chat translation evaluation, even surpassing \comet.

Hence, we complement our evaluation with an LLM-based fine-grained assessment of MT outputs derived using \textsc{ContextMQM} \citep{DBLP:journals/corr/abs-2403-08314}. The prompt includes the past eight bilingual source sentences as context and one in-domain in-context example with MQM assessment to elicit MQM-like evaluation from GPT-4o-mini\footnote{\texttt{gpt-4o-mini-2024-07-18} accessed on 10-2-2024.} for all systems submitted for the \de track.\footnote{Due to budget constraints, we conduct this evaluation only on \de, which had the highest number (eight) of participating teams.} Like \textsc{MQM}, we compute the segment-level \textit{error} score aggregating the number of minor, major, and critical errors, weighted by factors of 10, 5, and 1, respectively.

\section{Participants}
\label{sec:participants}

This section provides a brief description of each participant’s systems (\S~\ref{subsec:systems}). Table~\ref{tab:participants} summarizes details about the team's institutions and the language directions they participated in. Participants were asked to submit up to three systems per language direction: one primary (explicitly marked) and up to two contrastive systems. Next, we discuss the commonalities and differences between the different submissions \S~\ref{ssec:discussion}.

\subsection{Systems}
\label{subsec:systems}

\subsubsection{\baseline (Baseline)}
For our baseline model, we used the \baseline multilingual machine translation model \citep{DBLP:journals/corr/abs-2207-04672} based on an encoder-decoder Transformer architecture \citep{transformer-paper}. \baseline is trained to support over 200 languages, including those of interest in this shared task: English, German, French, Dutch, Brazilian Portuguese, and Korean.
We opted for a sentence-level baseline that does not incorporate additional context and used a beam size of 4 for generating translation hypotheses.

\subsubsection{\unbabel}
%\wafaa{one primary and two contrastive, not that these specifications for each team are per language pair.}
The joint submission of Unbabel and IT includes one primary submission and two contrastive submissions per language pair. The systems are based on Tower-7B models and are trained on the chat datasets released by the shared task. Their primary system uses contextual MBR re-ranking over a set of 50 candidates to get the best hypothesis. Additionally, the first contrastive submission is a 70B variant of the Tower model specialized to have general purpose translation capabilities and the second one uses greedy decoding with the 7B model finetuned on chat datasets.

\subsubsection{\deeptext}
%\wafaa{one primary}
\deeptext participated in the English-Korean language pair with a single primary system.
Their submission leverages Google's Gemma-2-27B model \footnote{\href{https://huggingface.co/google/gemma-2-27b-it}{\texttt{google/gemma-2-27b-it}}}, using the most recent two turns and summaries of previous turns as context, all within the same document. The turn summaries are generated using the GPT-4o-mini model. Their system was trained solely using the training data provided by the shared task.

\begin{table}[t]
    \centering
    \resizebox{\linewidth}{!}{
    \begin{tabular}{lp{4.5cm}p{2.5cm}}
    \toprule
      % \rowcolor{gray!24}
       \textbf{\textsc{Team}}  &   \textbf{\textsc{Institution}} &  \textbf{\textsc{Directions}} \\
       \midrule
\texttt{\deeptext} & Yonsei University & \ko \\
\texttt{\huwaei} & Huawei Translation Service Center & \de \\
\texttt{\multitan} & Université Paris Cité & \fr \\
\texttt{\adapt} & 
ADAPT research centre \& Dublin City University & \de, \fr \\
\texttt{\sheffield} & University of Sheffield & \de, \nl, \pt\\
\texttt{\clteam} & Vrije Universiteit Amsterdam &  \de, \nl, \fr, \pt \\
\texttt{\dcu} & Dublin City University & \textsc{all} \\
\texttt{\unbabel} &   Unbabel \&  Instituto de Telecomunicações &  \textsc{all} \\
% \midrule
   \cdashlinelr{1-3}
\texttt{Baseline} & Organizers & \textsc{all}\\
        \bottomrule
    \end{tabular}
    }
    \caption{The participating teams, their affiliations, and the language directions that they participated.}
    \label{tab:participants}
\end{table}

\subsubsection{\huwaei}
%\wafaa{one primary and two contrastive}
Huawei Translation Service Center (\huwaei) team submitted a primary and two contrastive systems for English$\leftrightarrow$German language pair. Their system is a 25-6 transformer encoder-decoder model with a feed-forward dimension of 4096 and 16 self-attention layers.
Their primary submission uses a model from the previous edition of the shared task as a baseline, finetuned on this edition’s training data, followed by a second finetuning on the validation data.
Next, they use MBR reranking to select the optimal candidate with \comet as the utility function using outputs generated from a diverse set of models. Their system then undergoes a self-training step on the MBR output. The contrastive submissions include models trained with different finetuning strategies (e.g. excluding the finetuning on the dev set). 

\subsubsection{\sheffield}
%\wafaa{one primary}
The \sheffield team participated in English$\leftrightarrow$German, English$\leftrightarrow$Dutch, and English$\leftrightarrow$Brazilian Portuguese, with one primary system per language pair. 
Their system performs low-rank \cite{hu2022lora} instruction-tuning with the training and validation datasets provided by the shared task on the Llama-3-8B-Instruct \footnote{\href{https://huggingface.co/meta-llama/Meta-Llama-3-8B-Instruct}{\texttt{meta-llama/Meta-Llama-3-8B-Instruct}}} model.
To incorporate contextual information and dependencies between chat messages, they introduce a context-aware sliding window approach that incorporates translations generated at each turn into the prompt. 

\begin{table*}[t]
\centering
 \def\arraystretch{1.1}
   \resizebox{\linewidth}{!}{
\begin{tabular}{llccccc}
\toprule
 \textbf{\textsc{Participant}}  & \textbf{\textsc{Base Model}}  &  \textbf{\textsc{Chat Context?}}  & \textbf{\textsc{In-domain Training?}}   & \textbf{\textsc{Multilingual?}} & \textbf{\textsc{Synthetic Data?}}  &  \textbf{\textsc{Decoding}} \\
 \midrule
\texttt{\deeptext} & Gemma-2-27B  & \cmark (summary)  & \cmark   & \xmark & \xmark & \textcolor{black!40}{NR}  \\
\texttt{\huwaei} & Transformer 25-6 & \xmark  & \cmark & \xmark &  \cmark  & MBR\\
% self training + back translation  \\
& (from scratch) \\
\texttt{\multitan} & Commercial* & \cmark & \cmark  & \xmark & \xmark & \textcolor{black!40}{NR} \\
\texttt{\adapt} & Llama-3-8B (\de) & \cmark (few-shot) & \cmark &  \xmark & \xmark  & \textcolor{black!40}{NR}\\
 &  NLLB-200-600M (\fr)  & \xmark & \cmark & \xmark & \cmark & \textcolor{black!40}{NR}\\
\texttt{\sheffield} & Llama-8b-Instruct & \cmark & \cmark   & \cmark & \xmark & \textcolor{black!40}{NR}\\
\texttt{\clteam} & TowerInstruct-7B-v0.2 &  \cmark (graph) & \xmark  & \cmark & \xmark & \textcolor{black!40}{NR} \\
\texttt{\dcu} & Llama3.1-8b & \textcolor{black!40}{NR} &  \cmark & \cmark & \xmark & \textcolor{black!40}{NR}\\
\texttt{\unbabel} & TowerBase-7B & \cmark  & \cmark & \cmark & \xmark & MBR \\
   \cdashlinelr{1-7}
\texttt{Baseline} & NLLB-3.3B & \xmark & \xmark  & \cmark & \xmark& Beam (4) \\
\bottomrule
 \end{tabular}}
\caption{Summary of approaches for all primary submissions. \textcolor{black!40}{NR}: Not reported. 
% \sweta{do others use greedy?}\wafaa{not reported.} 
} \label{tab:model_summary}

\end{table*}

\subsubsection{\adapt}
\adapt team submitted 3 (one primary and two contrastive) systems based on different pre-trained models: \textsc{NLLB}\footnote{\href{https://huggingface.co/facebook/nllb-200-distilled-600M}{\texttt{facebook/nllb-200-distilled-600M}}}, \textsc{mBART-50}\footnote{\href{https://huggingface.co/facebook/mbart-large-50-many-to-many-mmt}{\texttt{facebook/mbart-large-50-many-to-many-mmt}}} and Llama-3-8B\footnote{\href{https://huggingface.co/unsloth/llama-3-8b-bnb-4bit}{\texttt{unsloth/llama-3-8b-bnb-4bit}}}. Their primary system for \de uses a Llama-3-8B  backbone finetuned on the in-domain chat and a synthetic dataset generated by backtranslating domain-specific monolingual sentences. For \fr, they finetune an NLLB-600M model. During inference, with the LLM-based models, they perform few-shot prompting using examples retrieved via similarity search from the training dataset. Their contrastive systems are based on the encoder-decoder models but use the same datasets for training. 

\subsubsection{\multitan}
\multitan's primary system finetunes a ``Dialog'' in-domain specialized model hosted on the Model Studio Lite server \footnote{\href{https://modelstudio-lite.systran.net}{\texttt{modelstudio-lite}}}  with 2022 Chat Task (train, valid, test) and 2024 Chat Task (valid) datasets. Their two contrastive submissions use outputs from \baseline model and the Deep\_translator API respectively. All outputs are post-edited using GPT-4o.

\subsubsection{\dcu}
% \wafaa{one primary and two contrastive}
%% Ask them to confirm which contrastive submissions to use.

\dcu team submitted a total of 15 systems (one primary and two contrastive) for all the five language pairs. Their primary system finetunes a Llama-3.1-8B model on a mix of the chat task's training data and datasets from other WMT tracks. They also include synthetically generated customer-service data generated using one of their contrastive submission. Other contrastive submissions use Mistral-7B as base models with optional prompt tuning or finetuning of adapter layers. 

\subsubsection{\clteam}
% \wafaa{one primary and one contrastive}
\clteam submitted one primary and one contrastive systems for each of the English$\leftrightarrow$German, English$\leftrightarrow$French, English$\leftrightarrow$Dutch, and English$\leftrightarrow$Brazilian Portuguese language pairs.
Their system uses TowerInstruct-7B-v0.2 \footnote{\href{https://huggingface.co/Unbabel/TowerInstruct-7B-v0.2}{\texttt{Unbabel/TowerInstruct-7B-v0.2}}} model as the base LLM. For their primary submission, they prompt the model with both the dialogue history represented using a graph and the source sequence to be translated. To generate the graph, they prompt GPT-4o to extract entities and relationships from the dialogue data, creating triples from these elements. For the contrastive submission, they prompt the model with only the source sequence to be translated.

\subsection{Discussion} \label{ssec:discussion}
% \sweta{add summary of approaches here}
\Cref{tab:model_summary} presents a summary of approaches used by all the submitted systems. We highlight some key aspects below:

\paragraph{Model Architecture} Most teams except \clteam and \huwaei finetuned general-purpose pre-trained LLMs. Where \clteam used an off-the-shelf translation-finetuned LLM, \huwaei opted for a custom bilingual encoder-decoder model for their participation.

\begin{table*}[t]
    \centering
    \renewcommand{\arraystretch}{1.2}
    \footnotesize
    \resizebox{\linewidth}{!}{
    \begin{tabular}{lrrrrrrrrrr}
    \toprule
    &  \multicolumn{2}{c}{\textbf{\textsc{en-de}}} & \multicolumn{2}{c}{\textbf{\textsc{en-fr}}} & \multicolumn{2}{c}{\textbf{\textsc{en-nl}}} & \multicolumn{2}{c}{\textbf{\textsc{en-pt}}} & \multicolumn{2}{c}{\textbf{\textsc{en-ko}}}  \\
 
     \multirow{-2}{*}{\textbf{\textsc{System}}}  & \multicolumn{1}{c}{\textsc{de}} & \multicolumn{1}{c}{\textsc{en}} 
     & \multicolumn{1}{c}{\textsc{fr}} & \multicolumn{1}{c}{\textsc{en}} 
     & \multicolumn{1}{c}{\textsc{nl}} & \multicolumn{1}{c}{\textsc{en}}  
     & \multicolumn{1}{c}{\textsc{pt}} & \multicolumn{1}{c}{\textsc{en}}  
     & \multicolumn{1}{c}{\textsc{ko}} & \multicolumn{1}{c}{\textsc{en}}   \\
    \midrule   
\texttt{DeepText Lab}  & \cellcolor{gray!10} & \cellcolor{gray!10}& \cellcolor{gray!10} &\cellcolor{gray!10} & \cellcolor{gray!10} &\cellcolor{gray!10} & \cellcolor{gray!10} &\cellcolor{gray!10}  & 93.03 & 94.11 \\
\texttt{HW-TSC} & \textbf{93.58} & \textbf{93.30}  & \cellcolor{gray!10}& \cellcolor{gray!10}& \cellcolor{gray!10}& \cellcolor{gray!10}& \cellcolor{gray!10}& \cellcolor{gray!10} & \cellcolor{gray!10}& \cellcolor{gray!10}   
\\
\texttt{MULTITAN-GML}  &\cellcolor{gray!10} & \cellcolor{gray!10} & 90.09 & 92.42  & \cellcolor{gray!10}& \cellcolor{gray!10}& \cellcolor{gray!10}& \cellcolor{gray!10}& \cellcolor{gray!10}& \cellcolor{gray!10} \\
\texttt{ADAPT} 	 & 90.59 & 90.97  & 82.19 & 82.69 & \cellcolor{gray!10}& \cellcolor{gray!10}& \cellcolor{gray!10}& \cellcolor{gray!10}& \cellcolor{gray!10}& \cellcolor{gray!10}  \\
\texttt{SheffieldGATE} 	 & 88.67 & 90.10  & \cellcolor{gray!10}& \cellcolor{gray!10} & 88.93 & 89.71  & 90.05 & 88.12  & \cellcolor{gray!10}& \cellcolor{gray!10}
\\
\texttt{CLTeam}  & 90.90 & 91.63  & 91.37 & 91.90  & 91.31 & 91.22  & 91.77 & 90.12 & \cellcolor{gray!10}& \cellcolor{gray!10} \\
\texttt{DCUGenNLP}  & 90.49 & 91.10  & 91.05 & 90.73  & 91.32 & 90.96  & 93.24 & 89.66  & 91.50 & 93.41 \\
\texttt{Unbabel-IT}  & 93.22 & 92.48  & \textbf{92.96} & \textbf{92.71}  & \textbf{94.36} & \textbf{93.38}  & \textbf{94.76} & \textbf{92.46}  & \textbf{94.96} & \textbf{95.16}
  \\
  \midrule
  \texttt{\baseline}  & 90.56 & 89.03  & 91.06 & 89.18  & 87.86 & 88.45  & 86.33 & 86.10  & 87.26 & 88.05 \\
  $\Delta$ (Best) &  +3.02 & +4.27 & +1.9 & + 3.53 & +6.50 &	+4.93	& +8.43 &	+6.36 & +7.70 &	+7.11\\
    \bottomrule
    \end{tabular}}
    \caption{\comet results on the official test set. 
    % \textsc{xx} refer to language directions that translate into English and \textsc{en} refer to language directions that translate from English. 
    $\Delta$ (Best): improvement over baseline. }
    % \wafaa{add what \textsc{xx} and \textsc{en} refer to.} 
    \label{tab:comet_results}
\end{table*}

\begin{table*}[t]
    \centering
     \renewcommand{\arraystretch}{1.2}
    \footnotesize
    \resizebox{\linewidth}{!}{
    \begin{tabular}{lrrrrrrrrrr}
    \toprule
    &  \multicolumn{2}{c}{\textbf{\textsc{en-de}}} & \multicolumn{2}{c}{\textbf{\textsc{en-fr}}} & \multicolumn{2}{c}{\textbf{\textsc{en-nl}}} & \multicolumn{2}{c}{\textbf{\textsc{en-pt}}} & \multicolumn{2}{c}{\textbf{\textsc{en-ko}}}  \\
 
     \multirow{-2}{*}{\textbf{\textsc{System}}}   & \multicolumn{1}{c}{\textsc{de}} & \multicolumn{1}{c}{\textsc{en}} 
     & \multicolumn{1}{c}{\textsc{fr}} & \multicolumn{1}{c}{\textsc{en}} 
     & \multicolumn{1}{c}{\textsc{nl}} & \multicolumn{1}{c}{\textsc{en}}  
     & \multicolumn{1}{c}{\textsc{pt}} & \multicolumn{1}{c}{\textsc{en}}  
     & \multicolumn{1}{c}{\textsc{ko}} & \multicolumn{1}{c}{\textsc{en}}   \\
    \midrule   
\texttt{DeepText Lab} & \cellcolor{gray!10}& \cellcolor{gray!10} & \cellcolor{gray!10}& \cellcolor{gray!10} &\cellcolor{gray!10} & \cellcolor{gray!10} &\cellcolor{gray!10} & \cellcolor{gray!10} & 57.67 & 77.96 \\
\texttt{HW-TSC}  & \textbf{82.66} & \textbf{84.03} &  \cellcolor{gray!10}& \cellcolor{gray!10}& \cellcolor{gray!10}& \cellcolor{gray!10}& \cellcolor{gray!10}& \cellcolor{gray!10}   & \cellcolor{gray!10}& \cellcolor{gray!10}   
\\
\texttt{MULTITAN-GML} & \cellcolor{gray!10}& \cellcolor{gray!10} & 79.54 & \textbf{82.71} &  \cellcolor{gray!10}& \cellcolor{gray!10}& \cellcolor{gray!10}& \cellcolor{gray!10}& \cellcolor{gray!10}& \cellcolor{gray!10} \\
\texttt{ADAPT}  & 69.50 & 76.63  & 63.92 & 55.98  & \cellcolor{gray!10}& \cellcolor{gray!10}& \cellcolor{gray!10}& \cellcolor{gray!10}& \cellcolor{gray!10}& \cellcolor{gray!10}  \\
\texttt{SheffieldGATE}  & 64.94 & 72.04  & \cellcolor{gray!10}& \cellcolor{gray!10} & 60.01 & 68.31  & 67.67 & 66.38 & \cellcolor{gray!10}& \cellcolor{gray!10}
\\
\texttt{CLTeam} & 69.87 & 75.39  & 74.66 & 77.41  & 63.59 & 73.00  & 71.38 & 69.45  & \cellcolor{gray!10}& \cellcolor{gray!10} \\
\texttt{DCUGenNLP} & 69.84 & 73.64  & 73.73 & 73.78  & 67.44 & 70.47  & 75.24 & 67.27  & 49.02 & 75.35  \\
\texttt{Unbabel-IT} & 77.23 & 79.87  & \textbf{80.51} & 78.57  & \textbf{80.25} & \textbf{78.60}  & \textbf{82.55} & \textbf{76.01}  & \textbf{62.29} & \textbf{81.57} 
  \\
  \midrule
  \texttt{\baseline}  & 70.22 & 71.79  & 76.03 & 76.37  & 59.55 & 68.62  & 58.60 & 67.13  & 34.50 & 69.87 
 \\
 $\Delta$ (Best) & +12.44 &	+12.24 &	+4.48 &	+6.34 &	+20.70	& +9.98 &	+23.95 &	+8.88 &	+27.79 &	+11.70\\
    \bottomrule
    \end{tabular}}
    \caption{\chrf results on the official test set. $\Delta$ (Best): improvement over baseline.  }
    \label{tab:chrf_results}
\end{table*}

\paragraph{Training Data} All teams used the provided training and development data, sourced from the current and previous versions of the task. \huwaei went a step further by generating a synthetic parallel corpus. They did this by forward translating source-side monolingual data into target-side text and backtranslating target-side monolingual into source-side texts. \adapt similarly used back translation to generate more in-domain data for their \fr submission. 

\paragraph{Inference} Both \unbabel and \huwaei leveraged a quality-aware decoding (QAD) approach \cite{fernandes-etal-2022-quality} for further improving the quality of outputs during inference. While \huwaei optimized for \comet, \unbabel used a context-aware \comet metric as a utility for selecting the best candidate. \huwaei also used MBR outputs to further finetune the model.

\paragraph{Context Usage} Different strategies were employed to incorporate conversation context into the translation process. \unbabel, \sheffield, and \multitan utilized the previous turns of the conversation as context to maintain continuity and coherence in translations. \deeptext used both the previous two turns as well as the summary of all the previous conversation turns except the last two, generated by GPT-4o-mini. This allowed the model to focus on the essential part of the previous content without being overwhelmed by excessive details. On the other hand, \clteam used a graph representation of the conversation's history as context, capturing the connectivity between various concepts thus serving as a compressed memory of the dialogue context. \adapt used few shot examples extracted from the training data using sentence-embedding similarity.

All teams that participated for more than one language pair opted for a multilingual system except for \adapt team who submitted two different systems for each language pair they participated in (\de, \fr).

\begin{table*}[t]
    \centering
    \footnotesize
    \resizebox{\linewidth}{!}{
    \begin{tabular}{lrrrrrrrrrr}
    \toprule
    &  \multicolumn{2}{c}{\textbf{\textsc{en-de}}} & \multicolumn{2}{c}{\textbf{\textsc{en-fr}}} & \multicolumn{2}{c}{\textbf{\textsc{en-nl}}} & \multicolumn{2}{c}{\textbf{\textsc{en-pt}}} & \multicolumn{2}{c}{\textbf{\textsc{en-ko}}}  \\
 
     \multirow{-2}{*}{\textbf{\textsc{System}}}   & \multicolumn{1}{c}{\textsc{de}} & \multicolumn{1}{c}{\textsc{en}} 
     & \multicolumn{1}{c}{\textsc{fr}} & \multicolumn{1}{c}{\textsc{en}} 
     & \multicolumn{1}{c}{\textsc{nl}} & \multicolumn{1}{c}{\textsc{en}}  
     & \multicolumn{1}{c}{\textsc{pt}} & \multicolumn{1}{c}{\textsc{en}}  
     & \multicolumn{1}{c}{\textsc{ko}} & \multicolumn{1}{c}{\textsc{en}}   \\
    \midrule   
\texttt{DeepText Lab} &\cellcolor{gray!10} &  \cellcolor{gray!10}&\cellcolor{gray!10} & \cellcolor{gray!10} & \cellcolor{gray!10}& \cellcolor{gray!10} & \cellcolor{gray!10}&\cellcolor{gray!10}  & 15.99 & 16.15 \\
\texttt{HW-TSC}  & 20.79 & 23.37  & \cellcolor{gray!10}& \cellcolor{gray!10} & \cellcolor{gray!10}& \cellcolor{gray!10} & \cellcolor{gray!10}& \cellcolor{gray!10} & \cellcolor{gray!10}& \cellcolor{gray!10}
\\
\texttt{MULTITAN-GML} & \cellcolor{gray!10}& \cellcolor{gray!10} & 0.31 & 0.21  & \cellcolor{gray!10}& \cellcolor{gray!10}& \cellcolor{gray!10}& \cellcolor{gray!10}& \cellcolor{gray!10}& \cellcolor{gray!10} \\
\texttt{ADAPT}  & 15.63 & 17.97  & -23.31 & -22.88 & \cellcolor{gray!10}& \cellcolor{gray!10} & \cellcolor{gray!10}& \cellcolor{gray!10} & \cellcolor{gray!10}& \cellcolor{gray!10} \\
\texttt{SheffieldGATE} & 17.87 & 17.54  &\cellcolor{gray!10} & \cellcolor{gray!10} & 13.72 & 14.39  & 5.87 & 3.46& \cellcolor{gray!10}& \cellcolor{gray!10} 
\\
\texttt{CLTeam} & 19.65 & 21.15  & 8.22 & 7.26  & 19.00 & 19.20  & 8.64 & 7.68 & \cellcolor{gray!10}& \cellcolor{gray!10} \\
\texttt{DCUGenNLP}  & 17.27 & 20.38  & 5.11 & 4.80  & 16.55 & 16.09  & 8.69 & 6.70  & 15.84 & 15.74  \\
\texttt{Unbabel-IT}  & \textbf{24.41} & \textbf{26.15}  & \textbf{10.67} & \textbf{10.00}  & \textbf{23.93} & \textbf{23.39}  & \textbf{12.74} & \textbf{10.59}  & \textbf{21.64} & \textbf{21.08} 
  \\
  \midrule
  \texttt{\baseline} & 15.56 & 19.09  & 1.24 & 0.77  & 9.35 & 8.04  & -5.51 & -6.75  & 4.11 & 4.13  \\
   $\Delta$ (Best) & +8.85 & +7.06 & +9.43 & +9.23 & +14.58 & +15.35 & +18.25 & +17.34 & +17.53 & +16.95 \\
    \bottomrule
    \end{tabular}}
    \caption{\contextqe results on the official test set. $\Delta$ (Best): improvement over baseline. }
    \label{tab:cometqe_results}
\end{table*}

\begin{table*}[h]
    \centering
    \footnotesize
    \resizebox{\linewidth}{!}{
    \begin{tabular}{lrrrrrrrrrr}
    \toprule
    &  \multicolumn{2}{c}{\textbf{\textsc{en-de}}} & \multicolumn{2}{c}{\textbf{\textsc{en-fr}}} & \multicolumn{2}{c}{\textbf{\textsc{en-nl}}} & \multicolumn{2}{c}{\textbf{\textsc{en-pt}}} & \multicolumn{2}{c}{\textbf{\textsc{en-ko}}}  \\
 
     \multirow{-2}{*}{\textbf{\textsc{System}}}  & \multicolumn{1}{c}{\textsc{de}} & \multicolumn{1}{c}{\textsc{en}} 
     & \multicolumn{1}{c}{\textsc{fr}} & \multicolumn{1}{c}{\textsc{en}} 
     & \multicolumn{1}{c}{\textsc{nl}} & \multicolumn{1}{c}{\textsc{en}}  
     & \multicolumn{1}{c}{\textsc{pt}} & \multicolumn{1}{c}{\textsc{en}}  
     & \multicolumn{1}{c}{\textsc{ko}} & \multicolumn{1}{c}{\textsc{en}}  \\
    \midrule   
\texttt{DeepText Lab} & \cellcolor{gray!10}& \cellcolor{gray!10} & \cellcolor{gray!10}& \cellcolor{gray!10} &\cellcolor{gray!10} & \cellcolor{gray!10} &\cellcolor{gray!10} & \cellcolor{gray!10} & 37.65 & 66.98  \\
\texttt{HW-TSC} & \textbf{68.76} & \textbf{71.27} & \cellcolor{gray!10}& \cellcolor{gray!10} & \cellcolor{gray!10}& \cellcolor{gray!10} &\cellcolor{gray!10} & \cellcolor{gray!10} &\cellcolor{gray!10} & \cellcolor{gray!10}
\\
\texttt{MULTITAN-GML}  &\cellcolor{gray!10} & \cellcolor{gray!10} & 65.43 & \textbf{71.80} & \cellcolor{gray!10} &\cellcolor{gray!10}& \cellcolor{gray!10} &\cellcolor{gray!10}& \cellcolor{gray!10} &\cellcolor{gray!10} \\
\texttt{ADAPT} & 51.39 & 59.90  & 33.17 & 28.56  &\cellcolor{gray!10} & \cellcolor{gray!10} &\cellcolor{gray!10} & \cellcolor{gray!10} &\cellcolor{gray!10} & \cellcolor{gray!10} \\
\texttt{SheffieldGATE} & 41.15 & 50.72   &\cellcolor{gray!10} & \cellcolor{gray!10}& 33.62 & 46.54  & 42.58 & 42.25  &\cellcolor{gray!10} & \cellcolor{gray!10} 
\\
\texttt{CLTeam}& 50.41 & 55.71  & 57.05 & 61.09  & 39.29 & 55.41  & 46.42 & 49.34   &\cellcolor{gray!10} & \cellcolor{gray!10} \\
\texttt{DCUGenNLP} & 49.97 & 57.29  & 56.32 & 56.39  & 46.38 & 52.15  & 56.36 & 45.87  & 27.66 & 62.13  \\
\texttt{Unbabel-IT} & 61.45 & 62.86  & \textbf{66.41} & 63.18  & \textbf{65.70} & \textbf{63.75}  & \textbf{67.86} & \textbf{59.05}  & \textbf{41.54} & \textbf{71.01} 
  \\
  \midrule
  \texttt{\baseline}  & 50.43 & 52.09  & 59.21 & 58.07  & 33.55 & 48.47  & 28.25 & 45.59  & 12.46 & 49.76  \\
   $\Delta$ (Best) & +18.33	& +19.18 &	+7.20 &	+13.73 &	+32.15	& +15.28	& +39.61 &	+13.46 &	+29.08 &	+21.25 \\
    \bottomrule
    \end{tabular}}
    \caption{\bleu results on the official test set. $\Delta$ (Best): improvement over baseline. }
    \label{tab:bleu_results}
\end{table*}

\section{Overall Results} 

We present the results of the automatic evaluation for all participating systems for all language pairs in \S~\ref{ssec:automatic}. We then discuss findings from human evaluation in \S~\ref{ssec:human}, followed by an LLM-based error assessment of submitted systems for the \de task in \S~\ref{ssec:llm}.

\subsection{Automatic Evaluation} \label{ssec:automatic}

Tables~\ref{tab:comet_results}-\ref{tab:bleu_results} show the results of automatic evaluations on the official test set using \comet, \chrf \contextqe and \bleu respectively \---\ most participant systems improve the translation quality according to both neural (\comet, \contextqe) and lexical (\chrf, \bleu) metrics over the \baseline model, except the \adapt system for \fr. This can be explained by the fact that \adapt finetunes an NLLB-600M model for \fr, which, albeit from the same family of models as our baseline (\baseline), is significantly smaller in size.

The \unbabel submission consistently outperforms all other systems, except the \de translation task, where the winning submission according to \comet, \bleu, and \chrf is \huwaei. Similarly, \multitan scores the best on \bleu and \chrf when translating French into English. Interestingly both systems (\unbabel and \huwaei) use MBR decoding with \textsc{ContextComet} and \comet respectively, suggesting that inference optimization techniques like quality-aware decoding methods \cite{fernandes-etal-2022-quality} can be useful in pushing the translation quality of strong MT systems. 
However, as we will see in \S \ref{ssec:human}, this difference is not reflected in human assessments and in automatic metrics (\contextmqm and \contextqe), with different methods scoring the two systems differently. This highlights the importance of carefully selecting the optimized metrics and the evaluation criteria, as over-optimizing certain metrics may lead to mixed or misleading outcomes \cite{fernandes-etal-2022-quality}. 

\unbabel's submission also achieves the highest scores across all settings according to \contextqe. However, we observe that the range of quality scores produced by the \contextqe model, when aggregated at the system level, significantly deviates from the typical range of this metric.\footnote{System-level scores are higher when the context is not considered.} While \citet{DBLP:journals/corr/abs-2403-08314} demonstrate its effectiveness as a segment-level metric with improved correlation to human judgments, further investigation is necessary to understand how these system-level scores should be interpreted. For instance, \multitan, which performs well on lexical metrics such as \bleu and \chrf, receives a notably lower score with \contextqe.

\paragraph{Discourse Phenomena Analysis}

\begin{figure}[htb!]
\centering
\begin{subfigure}{0.92\linewidth}
  \centering
  \includegraphics[width=\textwidth]{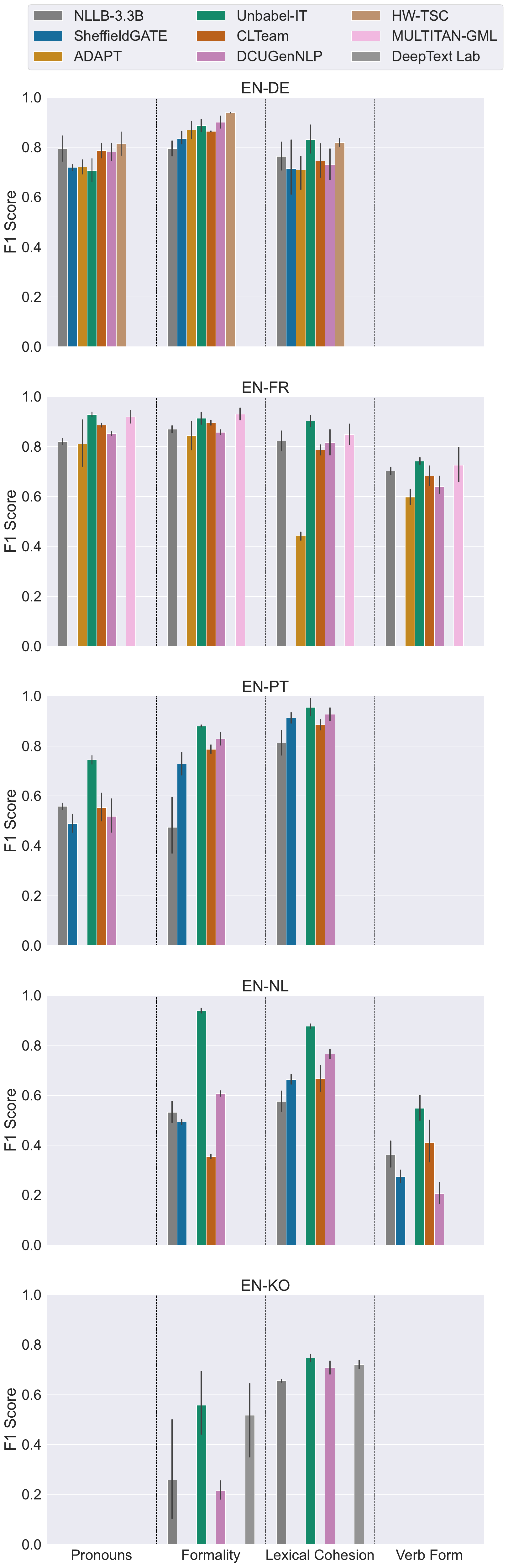}
\end{subfigure}%
\caption{\muda F1 scores across all settings.}
\label{fig:muda}
\end{figure}

% Muda
\begin{table}[htb!]
    \centering
    \resizebox{0.9\linewidth}{!}{
    \begin{tabular}{lccc}
    \toprule
        System & Precision & Recall & F1 \\
         \midrule
    \texttt{\huwaei}	 &76.7 &	86.2	 & 81.2 \\
\texttt{\adapt} & 75.0 & 69.2 & 72.0\\
\texttt{\sheffield}	&	73.0 &	70.8 &	71.9 \\
\texttt{\clteam}	&	75.7 &	81.5 &	78.5 \\
\texttt{\dcu}	&	74.6 &	81.5	& 77.9 \\
\texttt{\unbabel} &	75.4 &	66.2 &	70.5 \\
\midrule
\texttt{\baseline} & 74.3 &	84.6	& 79.1 \\
         \bottomrule
    \end{tabular}}
    \caption{\muda scores for \de pronouns.}
    \label{tab:muda_german_pronoun}
\end{table}

Figure~\ref{fig:muda} shows the F1 accuracy for all systems in correctly using the discourse markers across multiple phenomena for all language pairs. The baseline system (\baseline) has competitive accuracy with submitted systems on higher resource language pairs (\textsc{en$\rightarrow$de} and \textsc{en$\rightarrow$fr}). For all settings except ``pronouns'' for German and ``formality'' for German and French, \unbabel achieves the highest accuracy across the board. Surprisingly, the \muda F1 score for correctly generating German pronouns is worse for \unbabel relative to the baseline. A qualitative analysis shows that this is due to pronouns being under-generated in \unbabel's translations resulting in high precision but low recall scores as shown in Table~\ref{tab:muda_german_pronoun}.

To validate the observations and findings derived from automatic metrics, we now turn to human evaluation of the submitted systems for a more reliable assessment of translation quality.

% for placing

\begin{table*}[t]
    \centering
     \setlength{\tabcolsep}{2.5pt}
    \footnotesize
    \resizebox{0.95\linewidth}{!}{
    \begin{tabular}{lccccccccccccccc}
    \toprule
    &  \multicolumn{2}{c}{\textbf{\textsc{en-de}}} & \multicolumn{2}{c}{\textbf{\textsc{en-fr}}} & \multicolumn{2}{c}{\textbf{\textsc{en-nl}}} & \multicolumn{2}{c}{\textbf{\textsc{en-pt}}} & \multicolumn{2}{c}{\textbf{\textsc{en-ko}}}  \\
 
     \multirow{-2}{*}{\textbf{\textsc{System}}}  & \multicolumn{1}{c}{\textsc{xx}} & \multicolumn{1}{c}{\textsc{en}} & \multicolumn{1}{c}{\textsc{xx}} & \multicolumn{1}{c}{\textsc{en}} & \multicolumn{1}{c}{\textsc{xx}} & \multicolumn{1}{c}{\textsc{en}} & \multicolumn{1}{c}{\textsc{xx}} & \multicolumn{1}{c}{\textsc{en}} & \multicolumn{1}{c}{\textsc{xx}} & \multicolumn{1}{c}{\textsc{en}} \\
    \midrule   
\texttt{DeepText Lab} & \cellcolor{gray!10}& \cellcolor{gray!10} & \cellcolor{gray!10}& \cellcolor{gray!10} &\cellcolor{gray!10} & \cellcolor{gray!10} &\cellcolor{gray!10} & \cellcolor{gray!10} &91.35	 & 95.71  \\
\texttt{HW-TSC} & 88.47	 &90.41	& \cellcolor{gray!10}& \cellcolor{gray!10} & \cellcolor{gray!10}& \cellcolor{gray!10} & \cellcolor{gray!10}& \cellcolor{gray!10} & \cellcolor{gray!10}& \cellcolor{gray!10}
\\
\texttt{MULTITAN-GML} &  \cellcolor{gray!10}& \cellcolor{gray!10} & 81.83	& 84.62  & \cellcolor{gray!10}& \cellcolor{gray!10} & \cellcolor{gray!10}& \cellcolor{gray!10}  & \cellcolor{gray!10}& \cellcolor{gray!10}\\
\texttt{ADAPT} & 82.55 &	88.83 & 70.22 & 	65.53	  & \cellcolor{gray!10}& \cellcolor{gray!10} & \cellcolor{gray!10}& \cellcolor{gray!10} & \cellcolor{gray!10}& \cellcolor{gray!10} \\
\texttt{SheffieldGATE} & 78.63	 &88.85 & \cellcolor{gray!10}& \cellcolor{gray!10} &85.62	& 94.18  & 73.34	& 81.53  & \cellcolor{gray!10}& \cellcolor{gray!10}
\\
\texttt{CLTeam} & 83.12 &	89.12  & 84.28&	85.79	& 93.39 &	95.83  & 74.14 &	80.52 & \cellcolor{gray!10}& \cellcolor{gray!10}	  \\
\texttt{DCUGenNLP} & 84.56 &	88.60  & 85.72	&83.26 &  91.30 &	94.61 & 80.21 &	81.55 & 89.71 &	96.15 \\
\texttt{Unbabel-IT} & \textbf{89.42} &	\textbf{92.74}  & \textbf{90.24}	& \textbf{90.00}	& \textbf{98.16} &	\textbf{97.40}  & \textbf{82.04} &	\textbf{82.37} & \textbf{93.39} & \textbf{96.31} 
  \\
  \midrule
  \texttt{\baseline} &  78.05 &	87.57& 80.59	 & 77.82 & 82.66 &	90.98	 & 61.27 &	73.98  & 79.13 &	90.47 
 \\
    \bottomrule
    \end{tabular}
    }
    \caption{Human Evaluation results aggregated at the turn level on the official test set. }
    \label{tab:human_eval_results_turn}
\end{table*}

\begin{table}[t]
    \centering
     \setlength{\tabcolsep}{2.5pt}
    \renewcommand{\arraystretch}{1.2}
    \footnotesize
    \resizebox{\linewidth}{!}{
    \begin{tabular}{lccccc}
    \toprule
    &  \multicolumn{1}{c}{\textbf{\textsc{en-de}}} & \multicolumn{1}{c}{\textbf{\textsc{en-fr}}} & \multicolumn{1}{c}{\textbf{\textsc{en-nl}}} & \multicolumn{1}{c}{\textbf{\textsc{en-pt}}} & \multicolumn{1}{c}{\textbf{\textsc{en-ko}}}  \\

    \midrule   
\texttt{DeepText Lab}&\cellcolor{gray!10}&\cellcolor{gray!10}&\cellcolor{gray!10}&\cellcolor{gray!10}& 90.04 \\
\texttt{HW-TSC}	 &81.19& \cellcolor{gray!10}& \cellcolor{gray!10} & \cellcolor{gray!10}& \cellcolor{gray!10} 
\\
\texttt{MULTITAN-GML} &  \cellcolor{gray!10}& 	68.59 & \cellcolor{gray!10} & \cellcolor{gray!10} & \cellcolor{gray!10}\\
\texttt{ADAPT}	 &75.75	 & 59.65 & \cellcolor{gray!10} & \cellcolor{gray!10} & \cellcolor{gray!10} \\
\texttt{SheffieldGATE}  &	75.72 & \cellcolor{gray!10}&	70.81   &	68.27 & \cellcolor{gray!10}
\\
\texttt{CLTeam}  &	78.61 &73.32 &	84.37	& 69.85 & \cellcolor{gray!10} \\
\texttt{DCUGenNLP}  &	77.03  &	72.27&	76.41 &	73.78  &	89.83 \\
\texttt{Unbabel-IT} &	\textbf{84.22} 	& \textbf{79.62} &	\textbf{92.22}  &	\textbf{78.00} &	\textbf{93.21}
  \\
  \midrule
  \texttt{\baseline}  &	74.50  & 	67.81&  53.07 & 	56.37  &	85.63
 \\
    \bottomrule
    \end{tabular}}
    \caption{Human Evaluation results aggregated at the conversation level on the official test set. }
    \label{tab:human_eval_results_conv}
\end{table}
\begin{figure*}[t]
\centering
\begin{subfigure}{0.92\linewidth}
  \centering
\includegraphics[width=\textwidth]{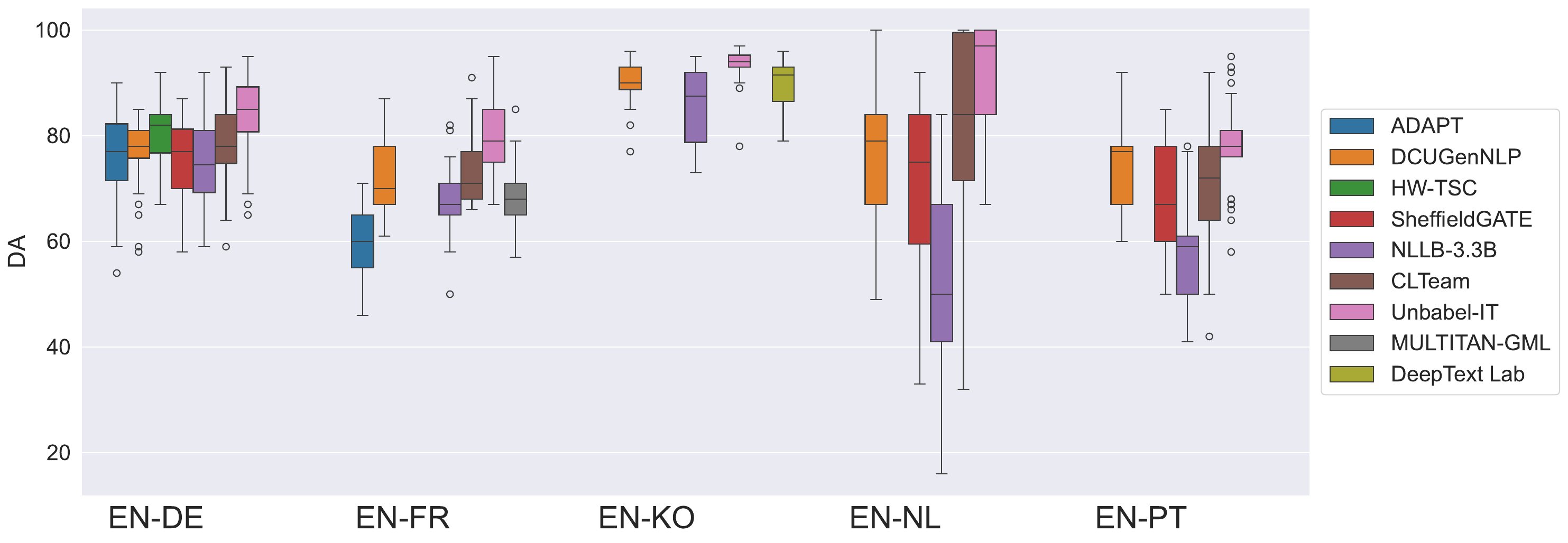}
\end{subfigure}%
 \caption{Conversation-level DA scores. }
 \label{fig:conv_sys_box}
\end{figure*}

\begin{figure*}[t]
    \centering
        \includegraphics[width=0.32\linewidth]{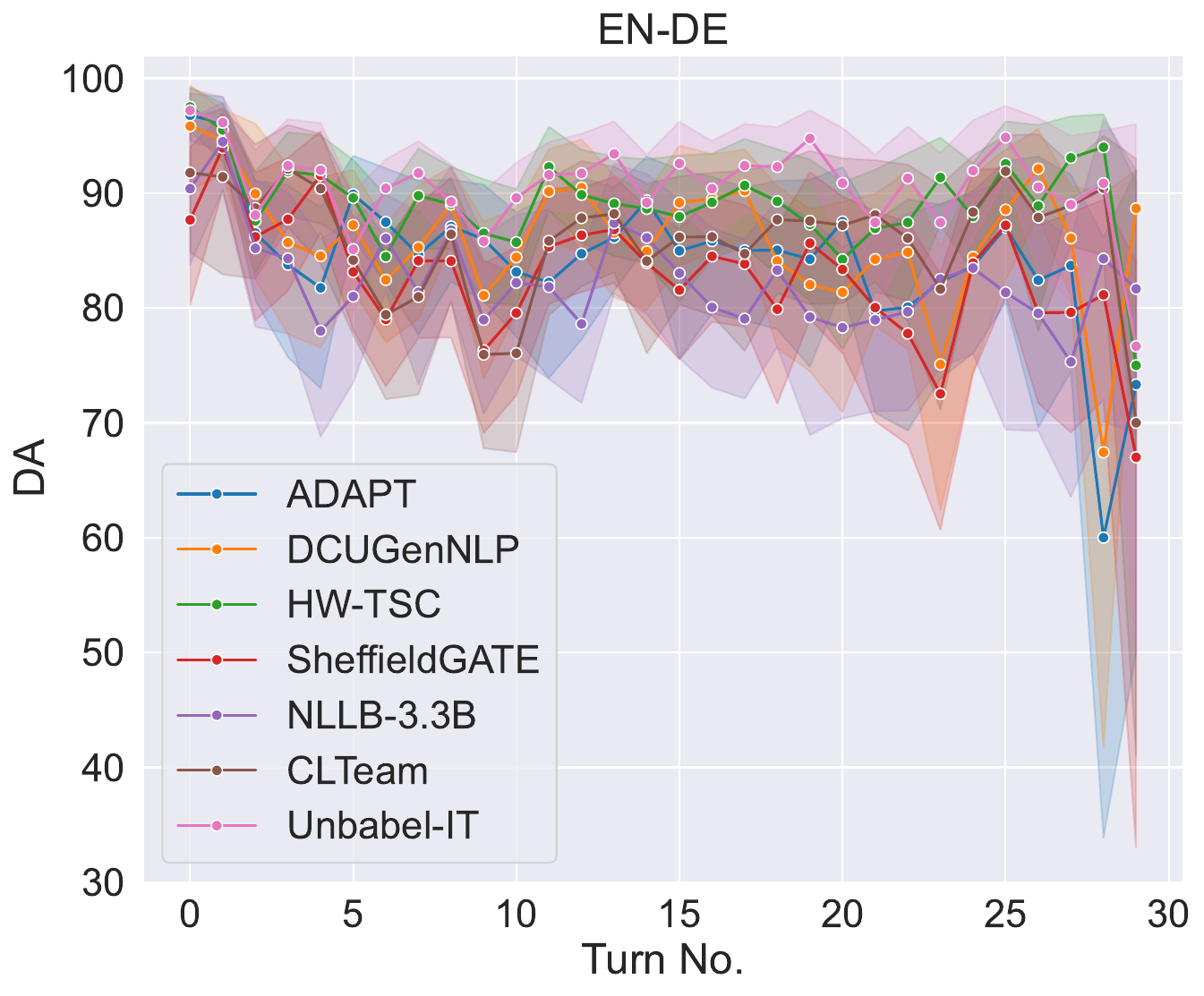}
        \includegraphics[width=0.32\linewidth]{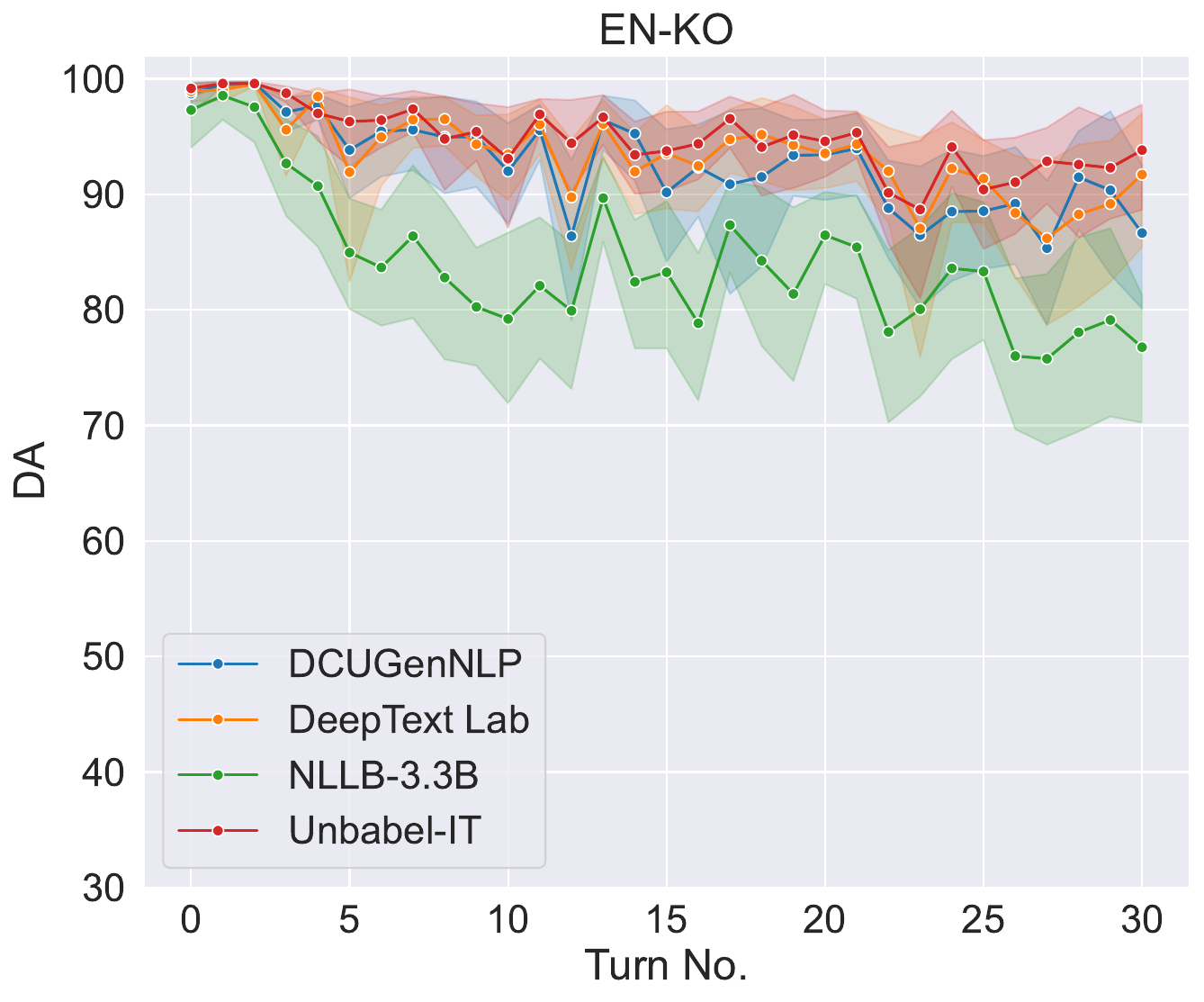}
        \includegraphics[width=0.32\linewidth]{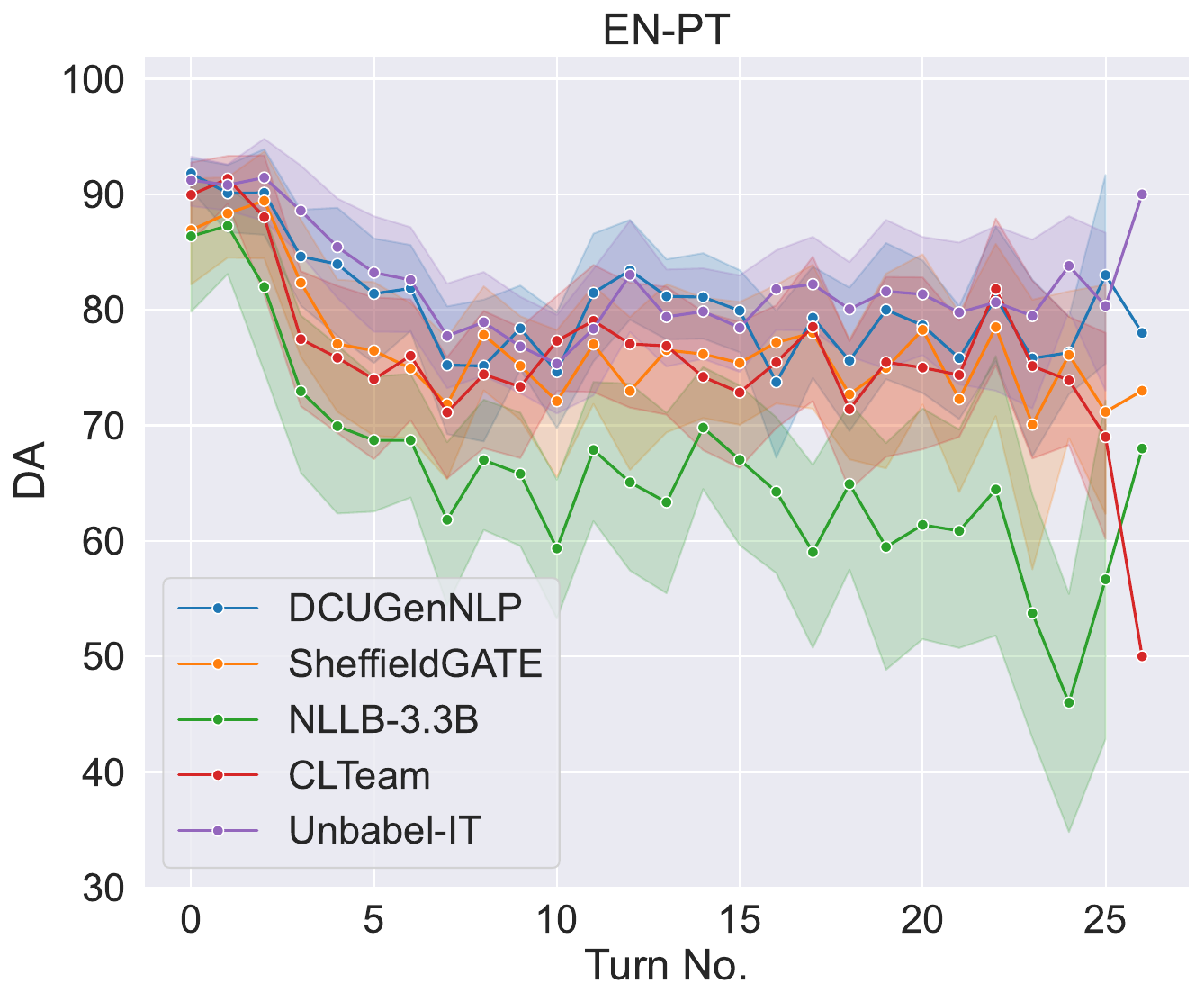}
        \includegraphics[width=0.32\linewidth]{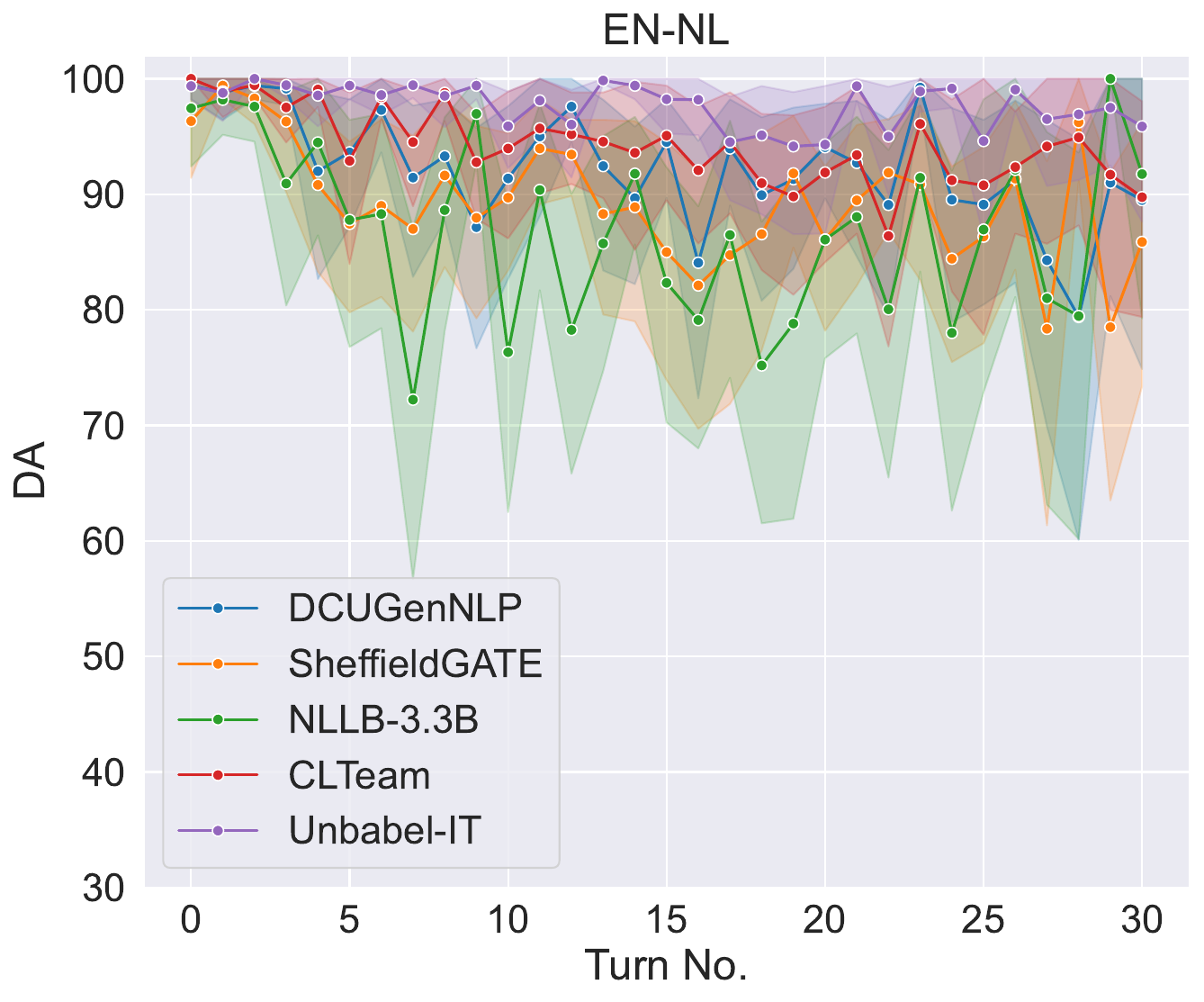}
        \includegraphics[width=0.32\linewidth]{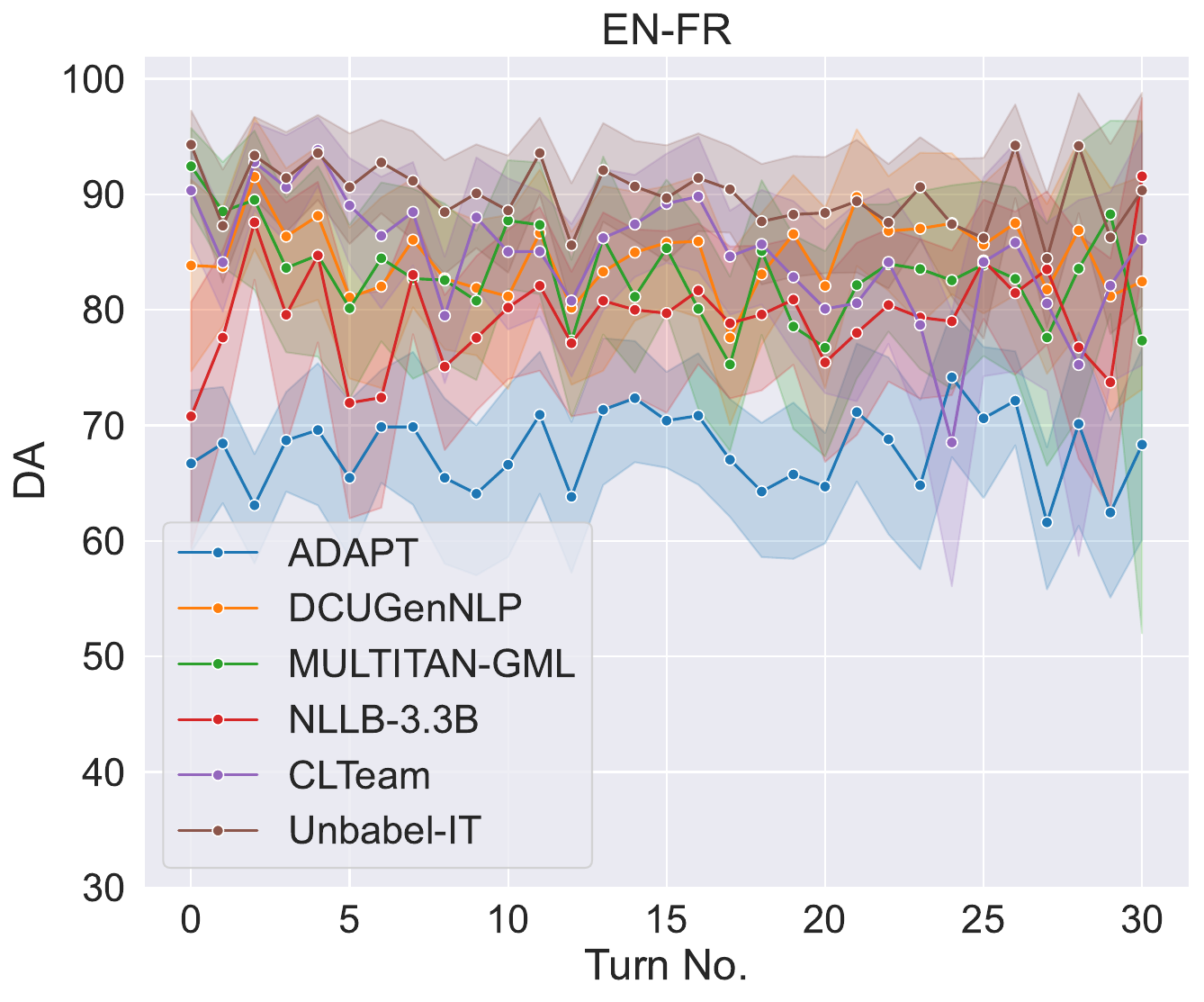}
    \caption{Turn-level DA score across different language pairs through a chat.}
    \label{fig:da_scores_turn}
\end{figure*}

\subsection{Human Evaluation} \label{ssec:human}

We present the human evaluation results at both turn and conversation levels in Tables \ref{tab:human_eval_results_turn} and \ref{tab:human_eval_results_conv} respectively. 

\paragraph{Overall results.} \unbabel outperforms all systems on both turn-level and conversation-level evaluation, surpassing the \huwaei system that achieved the highest \comet scores on \de translation pair.\footnote{We note that the human evaluation for \de, like other LPs, was conducted on a subset of the dataset (limited to a maximum of 30 turns per conversation).}
The translation quality according to direct assessment scores of all systems evaluated across all language pairs is high ($>$ 65) at both conversation and turn levels. This could be because of the nature of the chat dataset which contains very short texts (the number of words per turn across language pairs is less than 8, see Table~\ref{tab:data_stats}). 

\paragraph{Conversation-level results.} 
Figure~\ref{fig:conv_sys_box} shows the distribution of scores assigned at the conversation level for all systems and language pairs. Confirming the automatic results, \baseline scores the lowest and with the highest standard deviation for \ko, \nl and \pt. We also observe that \nl generally exhibits the largest standard deviation. After analyzing the outputs, we found that \nl has the highest number of segments (and conversations) receiving either a score of 0 (when hallucinating or copying source text verbatim) or 100, indicating a significant variation in translation quality for this language pair. Although there are sentences with mid-range scores, the dominance of segments with extremely high or low scores greatly influences the overall results, substantially raising the standard deviation.

\paragraph{Turn-level results.} 
Figure~\ref{fig:da_scores_turn} illustrates DA scores with the increase in the number of turns. For most systems and language pairs, translation quality deteriorates over successive turns, indicating a decline in the systems' ability to maintain consistency and accuracy in prolonged dialogues. This decline is particularly evident in the baseline system, which does not leverage contextual information from previous turns to generate translations. Interestingly, however, despite not using contextual information, \huwaei's system maintains translation quality across successive turns. This can likely be attributed to rigorous training on in-domain data, both authentic and synthetically generated.

\begin{figure}[t]
\centering
\begin{subfigure}{0.90\linewidth}
  \centering
\includegraphics[width=\textwidth]{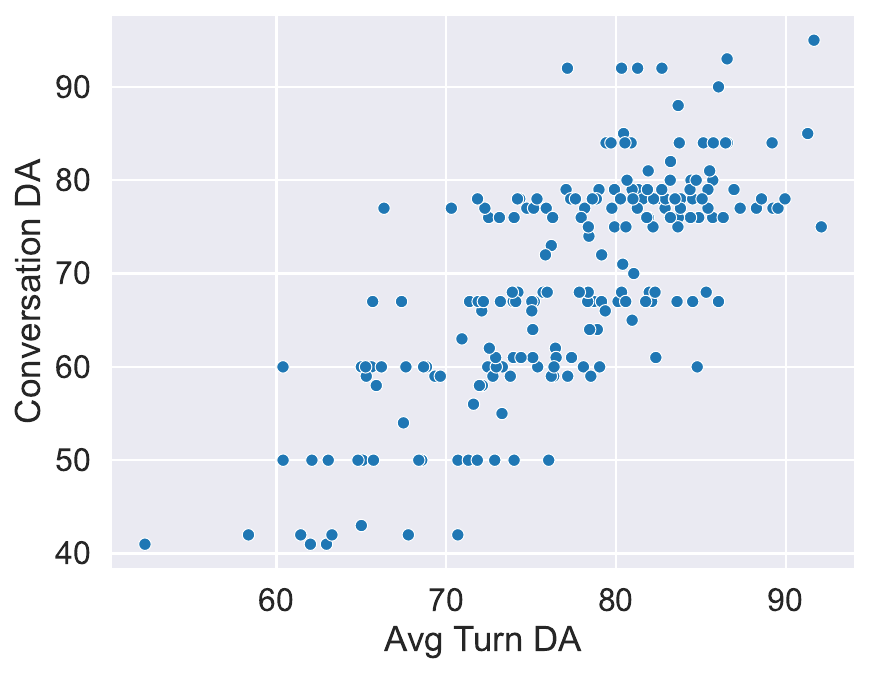}
\end{subfigure}%
 \caption{Turn avg. vs conversation-level DA scores.}
 \label{fig:turn_vs_conv}
\end{figure}

\paragraph{Turn Vs. Conversation Quality results.} Overall, \textbf{conversation-level quality is lower than turn-level scores} suggesting that there are aspects beyond translation accuracy that might impact the overall translation quality and user experience. This is corroborated by the observation that the Spearman correlation between the average turn-level score and conversation-level DA score, though high, is 0.722. For future evaluations, it might be worth investigating dialogue-oriented human assessment \cite{mendonca-etal-2023-dialogue} to understand how turn-level scores impact conversation-level quality. 

While direct assessments from experts provide a reliable measure of translation quality, DA scores fall short in offering insights into when and how errors occur, as well as their types and nature. Therefore, to assess the severity of errors generated by these systems, we now turn to LLM-based fine-grained error assessment of translation outputs.

\subsection{LLM-based Evaluation} \label{ssec:llm}

% ContextMQM
\begin{table}[h]
    \centering
    \setlength{\tabcolsep}{2.5pt}
    \resizebox{0.95\linewidth}{!}{
    \begin{tabular}{lcrrrc}
    \toprule
        \textbf{System} & \textbf{\% Perfect} & \textbf{\# Minor} & \textbf{\# Major} & \textbf{\# Critical} & \textbf{Avg. Score} \\
         \midrule
    \texttt{\huwaei} & 89.12 & 100 & 88 & 59 & -0.554 \\
\texttt{\adapt} & 82.61 & 158 & 139 & 99 & -0.903  \\
\texttt{\sheffield}	& 77.95 & 220 & 178 & 95 & -1.009 \\
\texttt{\clteam} & 86.28 & 139 & 82 & 79 & -0.656	 \\
\texttt{\dcu}	& 83.10 & 143 & 158 & 80 & -0.849\\
\texttt{\unbabel}  & \textbf{94.41} & \textbf{51} & \textbf{47} & \textbf{18} & \textbf{-0.228}	 \\
\midrule
\texttt{\baseline}  & 80.50 & 161 & 143 & 117 & -1.002  \\
         \bottomrule
    \end{tabular}}
    \caption{\contextmqm scores for \de.}
    \label{tab:contextmqm}
\end{table}

Table~\ref{tab:contextmqm} shows the results from using LLM-based error assessments via \contextmqm. \unbabel leads the pack with 94.41\% perfect translations. It also has the lowest number of errors in each category (minor, major, and critical), with an average error score of -0.228 (less than 1 minor error), the best among all systems. All systems, however, manage to achieve over 77\% perfect translations, meaning the overall quality across the board is strong. 

Despite the positive results, there are notable differences in error distribution. For example, both the \sheffield and \adapt models, while maintaining a reasonable percentage of perfect translations (82.61\% and 77.95\%, respectively), suffer from a significantly higher number of errors across all categories—minor, major, and critical. This suggests that when these systems do make errors, they tend to be more frequent and more serious, dragging down their overall performance compared to other systems.
Interestingly, contrary to human evaluation but in line with other automatic measures, \dcu scores worse than \clteam submission, highlighting limitations of existing evaluation methods to discern systems with close translation quality.

\section{Conclusions}

This paper presents the findings of the Chat Translation Shared Task 2024. This year, we expanded the set of language pairs to include two additional languages (\ko and \nl). We created the evaluation sets with a focus on context usage when assessing system performance. We also employed a range of complementary evaluation methods to assess all systems, including automatic metrics that focus on translation quality, as well as fine-grained error assessments and analysis of specific discourse phenomena.

We find that the best systems finetune strong pretrained LLMs using multilingual in-domain data and use contextual information (such as graphs, summaries or raw context) during training and inference. Additionally, using synthetic data during training improved translation quality. Furthermore, QAD strategies were effective in aligning translations with quality expectations.

As future work, a possible direction is to leverage reference-free discourse quality metrics that can give complementary insights to the translation evaluation approaches we tried this year. It might also be worth investigating human and automatic evaluation frameworks that assess specific dimensions relevant to chat (e.g. fluidity, coherence, consistency, etc).

\section*{Limitations}
Due to budget constraints, we conducted human evaluations using DA on a subset of the test set, which limited the number of turns evaluated for each language pair. For similar cost-related reasons, we ran \contextmqm on a single language pair that received the highest number of submissions. Additionally, we note that our analysis of discourse-specific phenomena is constrained by the quality of taggers, which only annotate specific properties based on predefined rules and may not fully capture all levels of ambiguity present in chat datasets. 

\section*{Ethics Statement}

\paragraph{Data released.} MAIA 2.0 corpus includes conversations from clients who gave written consent to use this data for research purposes as long as it follows the General Data Protection Regulation (GDPR). The original segments of customers and agents are translated via the MTPE (Machine Translation followed by a Post-Editing) process by experienced translators of the Unbabel Community. To make the data publicly available, the corpus was first anonymized automatically by using the Unbabel proprietary anonymization tool and then went through manual verification. 

\paragraph{Human evaluation.} Human direct assessment of system outputs was performed by professional translators hired via the UpWork platform and paid at professional rates.

\paragraph{Reproducibility.} For LLM-based error assessment, the continual updates of closed commercial models present challenges to the reproducibility of research. We have included the exact version of the model we used, along with the precise date of evaluation.

\section*{Acknowledgements}

This work was supported by the EU's Horizon Europe Research and Innovation Actions (UTTER, contract 101070631), by the project DECOLLAGE (ERC-2022-CoG 101088763), by the Portuguese Recovery and Resilience Plan through project C645008882-00000055 (Center for Responsible AI), and by Fundação para a Ciência e Tecnologia through contract UIDB/50008/2020.

\bibliography{anthology,custom}

\begin{thebibliography}{19}
\expandafter\ifx\csname natexlab\endcsname\relax\def\natexlab#1{#1}\fi

\bibitem[{A(2022)}]{a-2022-automating}
Sujan~Reddy A. 2022.
\newblock \href {https://doi.org/10.18653/v1/2022.naacl-srw.29} {Automating human evaluation of dialogue systems}.
\newblock In \emph{Proceedings of the 2022 Conference of the North American Chapter of the Association for Computational Linguistics: Human Language Technologies: Student Research Workshop}, pages 229--234, Hybrid: Seattle, Washington + Online. Association for Computational Linguistics.

\bibitem[{Agrawal et~al.(2024)Agrawal, Farajian, Fernandes, Rei, and Martins}]{DBLP:journals/corr/abs-2403-08314}
Sweta Agrawal, M.~Amin Farajian, Patrick Fernandes, Ricardo Rei, and Andr{\'{e}} F.~T. Martins. 2024.
\newblock \href {https://doi.org/10.48550/ARXIV.2403.08314} {Is context helpful for chat translation evaluation?}
\newblock \emph{CoRR}, abs/2403.08314.

\bibitem[{Costa{-}juss{\`{a}} et~al.(2022)Costa{-}juss{\`{a}}, Cross, {\c{C}}elebi, Elbayad, Heafield, Heffernan, Kalbassi, Lam, Licht, Maillard, Sun, Wang, Wenzek, Youngblood, Akula, Barrault, Gonzalez, Hansanti, Hoffman, Jarrett, Sadagopan, Rowe, Spruit, Tran, Andrews, Ayan, Bhosale, Edunov, Fan, Gao, Goswami, Guzm{\'{a}}n, Koehn, Mourachko, Ropers, Saleem, Schwenk, and Wang}]{DBLP:journals/corr/abs-2207-04672}
Marta~R. Costa{-}juss{\`{a}}, James Cross, Onur {\c{C}}elebi, Maha Elbayad, Kenneth Heafield, Kevin Heffernan, Elahe Kalbassi, Janice Lam, Daniel Licht, Jean Maillard, Anna~Y. Sun, Skyler Wang, Guillaume Wenzek, Al~Youngblood, Bapi Akula, Lo{\"{\i}}c Barrault, Gabriel~Mejia Gonzalez, Prangthip Hansanti, John Hoffman, Semarley Jarrett, Kaushik~Ram Sadagopan, Dirk Rowe, Shannon Spruit, Chau Tran, Pierre Andrews, Necip~Fazil Ayan, Shruti Bhosale, Sergey Edunov, Angela Fan, Cynthia Gao, Vedanuj Goswami, Francisco Guzm{\'{a}}n, Philipp Koehn, Alexandre Mourachko, Christophe Ropers, Safiyyah Saleem, Holger Schwenk, and Jeff Wang. 2022.
\newblock \href {https://doi.org/10.48550/ARXIV.2207.04672} {No language left behind: Scaling human-centered machine translation}.
\newblock \emph{CoRR}, abs/2207.04672.

\bibitem[{Deriu et~al.(2021)Deriu, Rodrigo, Otegi, Echegoyen, Rosset, Agirre, and Cieliebak}]{deriu2021survey}
Jan Deriu, Alvaro Rodrigo, Arantxa Otegi, Guillermo Echegoyen, Sophie Rosset, Eneko Agirre, and Mark Cieliebak. 2021.
\newblock Survey on evaluation methods for dialogue systems.
\newblock \emph{Artificial Intelligence Review}, 54:755--810.

\bibitem[{Farajian et~al.(2020)Farajian, Lopes, Martins, Maruf, and Haffari}]{farajian-etal-2020-findings}
M.~Amin Farajian, Ant{\'o}nio~V. Lopes, Andr{\'e} F.~T. Martins, Sameen Maruf, and Gholamreza Haffari. 2020.
\newblock \href {https://aclanthology.org/2020.wmt-1.3} {Findings of the {WMT} 2020 shared task on chat translation}.
\newblock In \emph{Proceedings of the Fifth Conference on Machine Translation}, pages 65--75, Online. Association for Computational Linguistics.

\bibitem[{Farinha et~al.(2022)Farinha, Farajian, Buchicchio, Fernandes, C.~de Souza, Moniz, and Martins}]{farinha-etal-2022-findings}
Ana~C Farinha, M.~Amin Farajian, Marianna Buchicchio, Patrick Fernandes, Jos{\'e}~G. C.~de Souza, Helena Moniz, and Andr{\'e} F.~T. Martins. 2022.
\newblock \href {https://aclanthology.org/2022.wmt-1.70} {Findings of the {WMT} 2022 shared task on chat translation}.
\newblock In \emph{Proceedings of the Seventh Conference on Machine Translation (WMT)}, pages 724--743, Abu Dhabi, United Arab Emirates (Hybrid). Association for Computational Linguistics.

\bibitem[{Federmann(2018)}]{federmann-2018-appraise}
Christian Federmann. 2018.
\newblock \href {https://aclanthology.org/C18-2019} {Appraise evaluation framework for machine translation}.
\newblock In \emph{Proceedings of the 27th International Conference on Computational Linguistics: System Demonstrations}, pages 86--88, Santa Fe, New Mexico. Association for Computational Linguistics.

\bibitem[{Fernandes et~al.(2023{\natexlab{a}})Fernandes, Deutsch, Finkelstein, Riley, Martins, Neubig, Garg, Clark, Freitag, and Firat}]{fernandes-etal-2023-devil}
Patrick Fernandes, Daniel Deutsch, Mara Finkelstein, Parker Riley, Andr{\'e} Martins, Graham Neubig, Ankush Garg, Jonathan Clark, Markus Freitag, and Orhan Firat. 2023{\natexlab{a}}.
\newblock \href {https://doi.org/10.18653/v1/2023.wmt-1.100} {The devil is in the errors: Leveraging large language models for fine-grained machine translation evaluation}.
\newblock In \emph{Proceedings of the Eighth Conference on Machine Translation}, pages 1066--1083, Singapore. Association for Computational Linguistics.

\bibitem[{Fernandes et~al.(2022)Fernandes, Farinhas, Rei, C.~de Souza, Ogayo, Neubig, and Martins}]{fernandes-etal-2022-quality}
Patrick Fernandes, Ant{\'o}nio Farinhas, Ricardo Rei, Jos{\'e}~G. C.~de Souza, Perez Ogayo, Graham Neubig, and Andre Martins. 2022.
\newblock \href {https://doi.org/10.18653/v1/2022.naacl-main.100} {Quality-aware decoding for neural machine translation}.
\newblock In \emph{Proceedings of the 2022 Conference of the North American Chapter of the Association for Computational Linguistics: Human Language Technologies}, pages 1396--1412, Seattle, United States. Association for Computational Linguistics.

\bibitem[{Fernandes et~al.(2023{\natexlab{b}})Fernandes, Yin, Liu, Martins, and Neubig}]{fernandes-etal-2023-translation}
Patrick Fernandes, Kayo Yin, Emmy Liu, Andr{\'e} Martins, and Graham Neubig. 2023{\natexlab{b}}.
\newblock \href {https://doi.org/10.18653/v1/2023.acl-long.36} {When does translation require context? a data-driven, multilingual exploration}.
\newblock In \emph{Proceedings of the 61st Annual Meeting of the Association for Computational Linguistics (Volume 1: Long Papers)}, pages 606--626, Toronto, Canada. Association for Computational Linguistics.

\bibitem[{Gon{\c{c}}alves et~al.(2022)Gon{\c{c}}alves, Buchicchio, Stewart, Moniz, and Lavie}]{goncalves-etal-2022-agent}
Madalena Gon{\c{c}}alves, Marianna Buchicchio, Craig Stewart, Helena Moniz, and Alon Lavie. 2022.
\newblock \href {https://aclanthology.org/2022.eamt-1.23} {Agent and user-generated content and its impact on customer support {MT}}.
\newblock In \emph{Proceedings of the 23rd Annual Conference of the European Association for Machine Translation}, pages 201--210, Ghent, Belgium. European Association for Machine Translation.

\bibitem[{Hu et~al.(2022)Hu, Shen, Wallis, Allen-Zhu, Li, Wang, Wang, and Chen}]{hu2022lora}
Edward~J Hu, Yelong Shen, Phillip Wallis, Zeyuan Allen-Zhu, Yuanzhi Li, Shean Wang, Lu~Wang, and Weizhu Chen. 2022.
\newblock Lora: Low-rank adaptation of large language models.
\newblock \emph{International Conference on Learning Representations}.

\bibitem[{Kocmi and Federmann(2023)}]{kocmi-federmann-2023-gemba}
Tom Kocmi and Christian Federmann. 2023.
\newblock \href {https://doi.org/10.18653/v1/2023.wmt-1.64} {{GEMBA}-{MQM}: Detecting translation quality error spans with {GPT}-4}.
\newblock In \emph{Proceedings of the Eighth Conference on Machine Translation}, pages 768--775, Singapore. Association for Computational Linguistics.

\bibitem[{Lu et~al.(2024)Lu, Qiu, Ding, Zhang, Kocmi, and Tao}]{lu-etal-2024-error}
Qingyu Lu, Baopu Qiu, Liang Ding, Kanjian Zhang, Tom Kocmi, and Dacheng Tao. 2024.
\newblock \href {https://doi.org/10.18653/v1/2024.findings-acl.520} {Error analysis prompting enables human-like translation evaluation in large language models}.
\newblock In \emph{Findings of the Association for Computational Linguistics ACL 2024}, pages 8801--8816, Bangkok, Thailand and virtual meeting. Association for Computational Linguistics.

\bibitem[{Mendonca et~al.(2023)Mendonca, Pereira, Menezes, Cabarr{\~a}o, Farinha, Moniz, Lavie, and Trancoso}]{mendonca-etal-2023-dialogue}
John Mendonca, Patr{\'\i}cia Pereira, Miguel Menezes, Vera Cabarr{\~a}o, Ana~C Farinha, Helena Moniz, Alon Lavie, and Isabel Trancoso. 2023.
\newblock \href {https://aclanthology.org/2023.gem-1.2} {Dialogue quality and emotion annotations for customer support conversations}.
\newblock In \emph{Proceedings of the Third Workshop on Natural Language Generation, Evaluation, and Metrics (GEM)}, pages 9--21, Singapore. Association for Computational Linguistics.

\bibitem[{Post(2018)}]{post-2018-call}
Matt Post. 2018.
\newblock \href {https://doi.org/10.18653/v1/W18-6319} {A call for clarity in reporting {BLEU} scores}.
\newblock In \emph{Proceedings of the Third Conference on Machine Translation: Research Papers}, pages 186--191, Brussels, Belgium. Association for Computational Linguistics.

\bibitem[{Rei et~al.(2022)Rei, C.~de Souza, Alves, Zerva, Farinha, Glushkova, Lavie, Coheur, and Martins}]{rei-etal-2022-comet}
Ricardo Rei, Jos{\'e}~G. C.~de Souza, Duarte Alves, Chrysoula Zerva, Ana~C Farinha, Taisiya Glushkova, Alon Lavie, Luisa Coheur, and Andr{\'e} F.~T. Martins. 2022.
\newblock \href {https://aclanthology.org/2022.wmt-1.52} {{COMET}-22: Unbabel-{IST} 2022 submission for the metrics shared task}.
\newblock In \emph{Proceedings of the Seventh Conference on Machine Translation (WMT)}, pages 578--585, Abu Dhabi, United Arab Emirates (Hybrid). Association for Computational Linguistics.

\bibitem[{Vaswani et~al.(2017)Vaswani, Shazeer, Parmar, Uszkoreit, Jones, Gomez, Kaiser, and Polosukhin}]{transformer-paper}
Ashish Vaswani, Noam Shazeer, Niki Parmar, Jakob Uszkoreit, Llion Jones, Aidan~N. Gomez, \L{}ukasz Kaiser, and Illia Polosukhin. 2017.
\newblock Attention is all you need.
\newblock In \emph{Proceedings of the 31st International Conference on Neural Information Processing Systems}, NIPS'17, page 6000–6010, Red Hook, NY, USA. Curran Associates Inc.

\bibitem[{Yeh et~al.(2021)Yeh, Eskenazi, and Mehri}]{yeh-etal-2021-comprehensive}
Yi-Ting Yeh, Maxine Eskenazi, and Shikib Mehri. 2021.
\newblock \href {https://doi.org/10.18653/v1/2021.eancs-1.3} {A comprehensive assessment of dialog evaluation metrics}.
\newblock In \emph{The First Workshop on Evaluations and Assessments of Neural Conversation Systems}, pages 15--33, Online. Association for Computational Linguistics.

\end{thebibliography}
\bibliographystyle{acl_natbib}

\end{document}